\PassOptionsToPackage{dvipsnames,HTML}{xcolor}
\documentclass{article} 
\usepackage{iclr2027_conference}
\usepackage{times}

\usepackage{amsmath,amsfonts,bm}

\def\Figref#1{Figure~\ref{#1}}

\def\eqref#1{equation~\ref{#1}}

\def\1{\bm{1}}

\DeclareMathAlphabet{\mathsfit}{\encodingdefault}{\sfdefault}{m}{sl}
\SetMathAlphabet{\mathsfit}{bold}{\encodingdefault}{\sfdefault}{bx}{n}

\newcommand{\E}{\mathbb{E}}

\usepackage[breaklinks=true]{hyperref}
\usepackage{url}
\usepackage{subcaption}
\usepackage{graphicx}
\usepackage{placeins}
\usepackage{afterpage}
\usepackage{bbold}
\usepackage[most]{tcolorbox}
\usepackage{xcolor}
\definecolor{borderblue}{HTML}{34495E}

\title{Diagnosing Visual Ignorance in Vision-Language Models}

\author{
    {\bf Runyu Zhou}\qquad
    {\bf Qi Zhang}\qquad
    {\bf Yisen Wang}\thanks{Corresponding author: Yisen Wang (yisen.wang@pku.edu.cn)} \\
    \ Peking University
}

\newcommand{\qwenthreeb}{Qwen2.5-VL-3B-Instruct}
\newcommand{\qwensevenb}{Qwen2.5-VL-7B-Instruct}
\newcommand{\llava}{LLaVA-v1.6-Mistral-7B}
\newcommand{\qwenthreebshort}{Qwen-3B}
\newcommand{\qwensevenbshort}{Qwen-7B}
\newcommand{\llavashort}{LLaVA-7B}

\newcommand{\ds}{\mathcal{D}}
\newcommand{\ksz}[1]{#1\times #1} 
\newcommand{\countrange}{$\{1,\ldots,9\}\}$}
\newcommand{\kersizes}{\ksz{1}, \ksz{3}, \ksz{5}, \ksz{7}, \ksz{11}, \ksz{15}, \ksz{31}, \ksz{61}}
\newcommand{\dseqprob}{\bar{E}_{\ds,k_i}}
\newcommand{\dsconsprob}{\bar{C}_{\ds,k_i}}
\newcommand{\dsbluracc}{\bar{\alpha}_{\ds,k_i}}
\newcommand{\dsconsacc}{\bar{\gamma}_{\ds,k_i}}

\iclrfinalcopy
\begin{document}

\maketitle

\begin{abstract}
Vision-language models (VLMs) achieve high accuracy on many visual question-answering benchmarks, yet it remains unclear whether this accuracy reflects reliable use of the visual details a question depends on. We examine this tension through the lens of visual ignorance: cases where a model can represent task-relevant visual evidence without using it to determine the answer. Across three VLMs and twelve benchmarks, we track the answer-relevant information decodable at each decoder layer and measure when it becomes coupled to the final output. The visual evidence needed for the correct answer can already be read out from intermediate layers, yet it has little effect on the output until the final layers of the decoder. Causal residual interchange confirms that shifting the answer requires the late states, not the earlier ones. We then evaluate all twelve benchmarks under progressive Gaussian blur and find that substantial fractions of predictions survive the entire degradation sequence while continuing to receive benchmark credit. These results distinguish the availability of visual information from its influence on prediction, and show that standard accuracy can coexist with limited sensitivity to the tested visual detail. They motivate training and evaluation that make answer-relevant visual distinctions necessary to answer correctly.
\end{abstract}

\section{Introduction}
Vision-language models (VLMs) combine visual encoders with language decoders pretrained on large text corpora~\citep{llava,instructblip,qwen25,vilp}. They can recognize familiar image content while overlooking details that determine the answer, particularly when those details conflict with learned expectations~\citep{langprior2023,vlindbench,blindfaithtext,vlmbias,mirage}. This phenomenon, called visual ignorance, has been documented behaviorally, but its internal basis and its consequences for evaluation remain unclear. We examine where task-relevant visual evidence becomes associated with and influences answer revision, and whether standard VQA benchmarks penalize insensitivity to that evidence.

\afterpage{%
\begin{figure}[t]
    \centering
    \includegraphics[width=\linewidth]{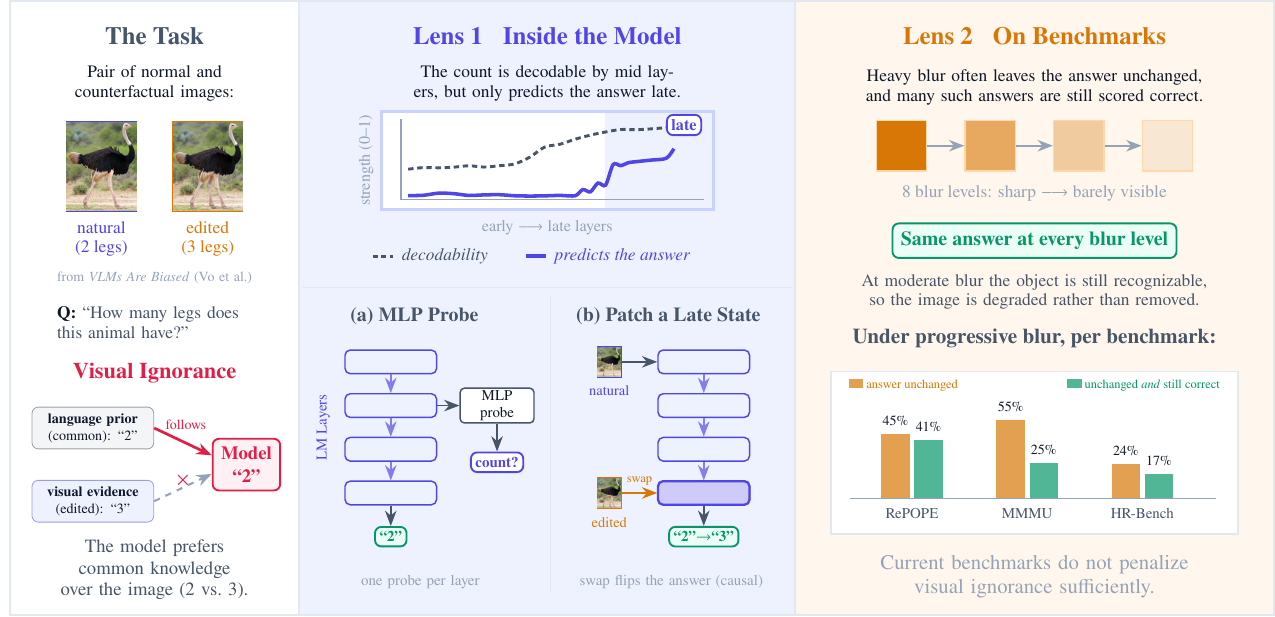}
    \caption{\textbf{Two lenses on visual ignorance in VLMs.} On paired
    normal/counterfactual images~\citep{vlmbias}, the model returns the prior count
    (``2'') for both. \emph{Lens~1:} per-layer MLP probes show the count is
    decodable at intermediate layers but strongly coupled to output changes only late;
    replacing a late state shifts the answer margin (\qwenthreeb; replicated observationally in 7B and
    LLaVA). \emph{Lens~2:} under progressive blur many answers stay unchanged and
    still score correct, so current benchmarks do not penalize visual ignorance
    sufficiently.}
    \label{fig:overview}
\end{figure}
}

We study these questions from two directions. Internally, we use matched image pairs in which a visual edit changes the correct answer while the question stays the same. Across \qwenthreeb{} (\qwenthreebshort)~\citep{qwen25}, \qwensevenb{} (\qwensevenbshort), and \llava{} (\llavashort)~\citep{llava}, we ask whether the internal update induced by the edit at each layer predicts the answer in the output. We find that in many pairs a trained probe can recover the edited answer from intermediate-layer states, yet these intermediate representations are only weakly coupled to the output; the association becomes strong only in the final layers of the decoder, where late readouts often favor the answer expected from learned priors rather than the edited visual evidence. In \qwenthreebshort{}, causal residual interchange confirms that it is the late states, not the earlier ones, that move the answer margin, and that within the sharpest transition the effect is attributable mainly to attention. The coupling pattern recurs in all three models, though the depth at which it begins varies.

Externally, we evaluate twelve VQA benchmarks under progressive Gaussian blur. For each prediction we track whether the answer stays fixed through every level of degradation, rather than changing and later returning to the original answer. On several benchmarks a large fraction of predictions survive the entire blur sequence. On RePOPE, for example, about half remain unchanged through the strongest blur, and this invariant subset stays above 90\% accuracy even as overall accuracy falls sharply. Inspecting these invariant answers reveals the same tendency observed inside the decoder: as the image is degraded, some explanations drift from visual description toward familiar associations, retaining the prior-consistent answer even when it is wrong. These results show that benchmark accuracy can stay high among predictions that do not respond to the degraded detail, revealing a gap between reported performance and use of the visual evidence a benchmark is intended to test.

This paper makes two contributions. First, we provide a causal, mechanistic account of visual ignorance: paired readouts and residual interchange show that answer-relevant evidence is often decodable in intermediate layers but shapes the output only late in the decoder, which tends to favor a learned prior; in \qwenthreebshort{} this late shift is driven mainly by attention, and the pattern holds across three VLMs. Second, we propose a degradation-based evaluation that blurs images progressively under a consecutive-agreement criterion across twelve benchmarks, quantifying how much credit goes to blur-insensitive predictions. Together these results separate the availability of visual information from its influence on prediction, and motivate training and evaluation that make answer-relevant visual distinctions necessary to answer correctly.

\section{Related Work}

\paragraph{Behavioral tests of visual dependence.}
Prior work documents VLM outputs predictable from text alone~\citep{langprior2023,vlindbench,vilp}, preference for textual cues when text and images conflict~\citep{blindfaithtext}, and stability after image removal~\citep{mirage}, showing that high accuracy need not imply dependence on instance-specific evidence. Yet an answer surviving image removal may reflect chance in a constrained answer space. We instead degrade the image gradually and require the answer to stay fixed across the whole sequence, reducing coincidental matches. Accuracy within this invariant subset then exposes benchmarks that credit predictions insensitive to the degraded detail.

\paragraph{Counterfactual and mechanistic analyses of visual information use.}
Counterfactual and conflict-based studies examine competition between visual evidence and learned priors. Visual CounterFact uses contrastive images to study pixel--prior competition and construct activation-steering directions~\citep{pixelsvspriors}; VLMBias tests whether familiar expectations override edited visual evidence~\citep{vlmbias}; and related analyses show that visual detail can be preserved without guiding the answer~\citep{vlmblind,hiddenplainsight}. Mechanistic studies use vocabulary projections, trained lenses, sparse features, attention analyses, and interventions to examine representations and information flow~\citep{logitlens,tunedlens,sae,saellm,saevlm,devils,interpreting}. VLM studies further investigate visual-information transfer, knowledge evolution, and modality arbitration~\citep{whatsinimg,knowledgeevolve,arbitration}. These methods establish such gaps observationally, associating representations with outputs but not testing whether the implicated states cause the answer. We instead use matched VLMBias pairs, where a normal and a counterfactual image share the question but differ in the correct count. We measure whether the internal update between the two images predicts the output change, and swap these states between paired images to reveal a sharp late-layer transition in their causal influence on the answer margin.

\section{Diagnosing Internal Visual Updates}
\label{sec:internal}

A VLM exhibits visual ignorance when an answer-relevant image change fails to produce an appropriate prediction update. We analyze this failure in three stages. First, layer replacement provides an initial localization, while per-layer semantic probes track where answer information is decodable. Second, matched normal--counterfactual pairs, in which the visual edit changes the correct answer, reveal when image-induced semantic updates become coupled to output changes, and residual interchange tests their causal influence on the output margin. Third, we test whether the coupling pattern generalizes across architectures. A consistent picture emerges: information about the answer may already be decodable from intermediate states, but image-induced updates remain weakly coupled to the output until a sharp late-layer transition. This late-coupling pattern recurs across architectures. In \qwenthreebshort{}, the sharpest transition is causally localized to attention in layer 27.

\subsection{Coarse Layer Localization}

We begin with coarse layer localization in a \qwenthreebshort{} model fine-tuned on VLMBias counterfactual pairs. Each pair contains a normal image and an edited counterpart with the same question but a different correct answer. For example, a two-legged ostrich and an edited three-legged ostrich are paired with ``How many legs does this animal have?'' Fine-tuning increases accuracy on the analyzed subset from 11.6\% to 65.2\%. We replace contiguous groups of baseline decoder layers with their fine-tuned counterparts and measure the resulting improvement. Accuracy rises from 11.6\% to 65.2\% as replacement expands, with the largest increase across the final 14--20 layers; replacing only the middle layers raises accuracy to 24.3\%. These results provide an initial localization to the middle and late decoder. Because hybrid networks can introduce representation mismatch, we next examine the frozen model layer by layer. Training details and examples appear in Appendix~\ref{app:coarse-internal}.

\subsection{Per-Layer Semantic Readout}

\begin{figure}[h]
    \centering
    \begin{subfigure}[b]{0.32\linewidth}
        \includegraphics[width=\linewidth,trim=0 12bp 0 12bp,clip]{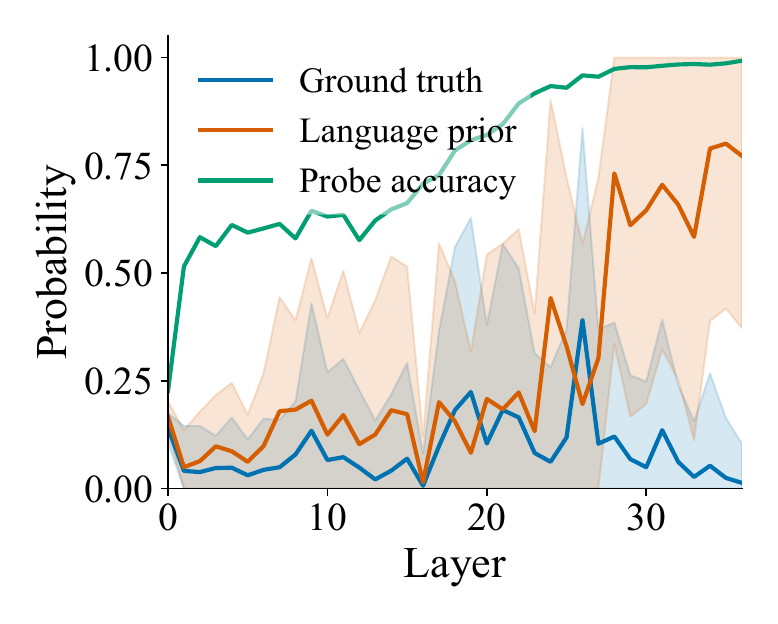}
        \caption{\qwenthreeb.}
        \label{fig:full-softmax-qwen3b}
    \end{subfigure}
    \hfill
    \begin{subfigure}[b]{0.32\linewidth}
        \includegraphics[width=\linewidth,trim=0 12bp 0 12bp,clip]{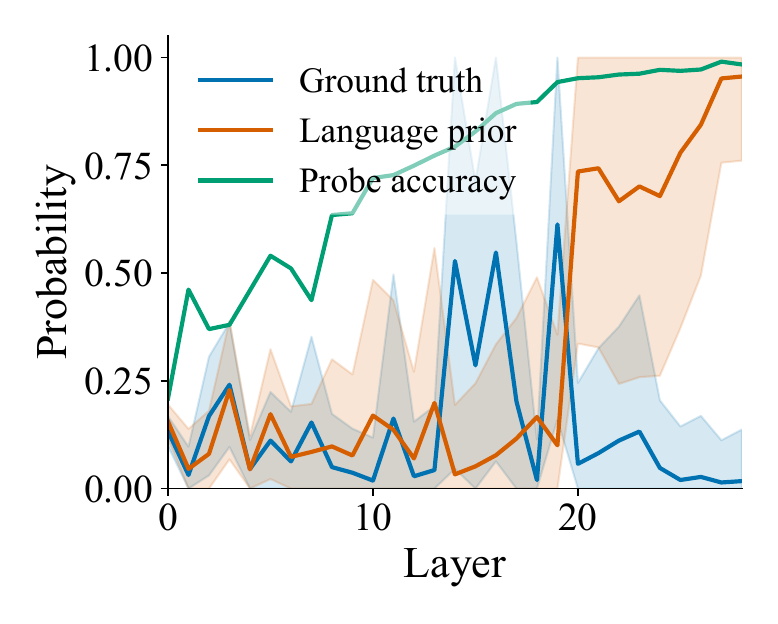}
        \caption{\qwensevenb.}
        \label{fig:full-softmax-qwen7b}
    \end{subfigure}
    \hfill
    \begin{subfigure}[b]{0.32\linewidth}
        \includegraphics[width=\linewidth,trim=0 12bp 0 12bp,clip]{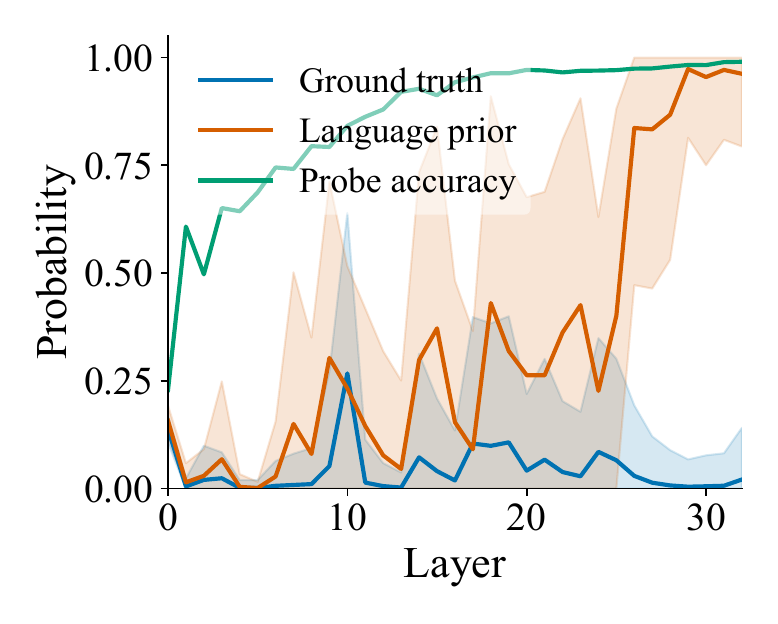}
        \caption{\llava.}
        \label{fig:full-softmax-llava}
    \end{subfigure}
    \caption{Full-softmax semantic probe readout. At every decoder layer, an independently trained nine-class MLP assigns probability to all counts in \countrange{}; blue and orange are the mean probabilities of the GT and LP counts on the filtered VLMBias evaluation subset, with standard-deviation bands. Green is held-out probe accuracy.}
    \label{fig:full-softmax-readout}
\end{figure}

Within each frozen model, we use independently trained MLP probes to read answer-related information from the final prompt-token state at every decoder layer. We focus on VLMBias examples whose correct answers are numerals from 1 to 9, and each probe predicts a distribution over these nine candidate counts; training details appear in Appendix~\ref{app:semantic-readout}. For each counterfactual image, we track the probability assigned to the visual ground-truth count (GT) and the familiar count favored by the language prior (LP). \Figref{fig:full-softmax-readout} shows that both counts are decodable across multiple layers, while late readouts increasingly favor LP. More directly, among Qwen examples whose final answers follow LP, GT receives over 50\% probe probability at an intermediate layer with over 90\% held-out probe accuracy in 49.8\% of 3B cases and 62.5\% of 7B cases, before LP becomes persistently dominant. Thus, in many failures, the visual ground-truth answer is strongly decodable before later states favor the language prior. We next use matched image pairs to test whether these internal changes are coupled to changes in the output.

\subsection{Paired Counterfactual Updates}
\label{sec:paired-counterfactual-updates}

Single-image readouts show which answers are decodable at each layer, but not whether a visual edit changes those representations or reaches the output. We therefore analyze 197 matched VLMBias pairs, each holding the question fixed while an image edit changes the correct count from the familiar count $A_{\mathrm{LP}}$ to the visual ground-truth count $A_{\mathrm{GT}}$. Pairwise comparison removes shared preferences and isolates the response to the edit. We use two matched wordings: Q1 retains the dataset form, e.g., ``How many legs does this animal have?'', while Q2 uses ``Count the legs of this animal.'' Both require a single numeral. Under Q1, \qwenthreebshort{} answers 138 normal images correctly but only 6 counterfactual images, retaining the original answer in 131 of the 138 initially correct cases. Q2 reproduces this failure in 120 of 132 cases. To link this failure to internal changes, we reuse the readout probes. For pair $p$ at layer $l$, the probe margin $s_{p,l}$ is the logit assigned to $A_{\mathrm{GT}}$ minus that assigned to $A_{\mathrm{LP}}$; the output margin $t_p$ uses the model's logits for the same counts. We define:
\[
u_{p,l}=s_{p,l}^{\mathrm{cf}}-s_{p,l}^{\mathrm{normal}},\qquad
M_p=t_p^{\mathrm{cf}}-t_p^{\mathrm{normal}},
\]
where cf denotes the counterfactual image. Positive values indicate movement toward the edited count. If an internal semantic update contributes to answer revision, larger $u_{p,l}$ should predict larger $M_p$.

\subsection{Internal--Output Update Coupling}
\begin{figure}[h]
    \centering
    \includegraphics[width=0.98\linewidth,trim=0 10bp 0 10bp,clip]{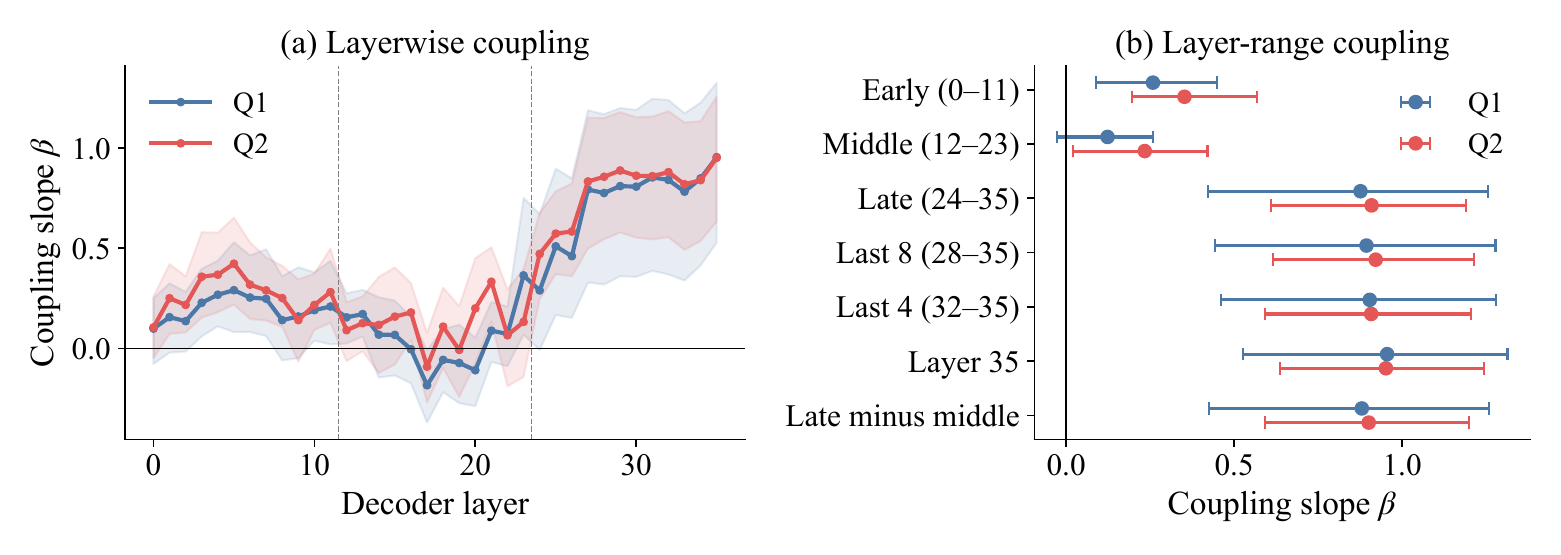}
    \caption{Internal--output update coupling in \qwenthreeb{}. (a) Layerwise coupling slopes under two prompts Q1 and Q2. (b) Coupling for layer-range averages and the late-minus-middle contrast. Shading and bars show 95\% confidence intervals.}
    \label{fig:paired-bridge}
\end{figure}

We quantify this relationship by regressing the output update $M_p$ on the internal update $u_{p,l}$ at each layer. The slope $\beta$ measures output-margin change per standard deviation of internal update. \Figref{fig:paired-bridge}(a) reports the layerwise slopes under both question wordings, while panel (b) reports coupling for layer-range averages and the late-minus-middle contrast. Full experimental and statistical details appear in Appendix~\ref{app:paired-bridge-details}.

As \Figref{fig:paired-bridge} shows, coupling remains weak through layers 12--23, where Q1 yields $\beta=0.123$, and then jumps sharply after layer 24. Over layers 24--35, $\beta$ reaches $0.876$ for Q1 and $0.909$ for Q2. Prior work shows that visual information and multimodal knowledge evolve across decoder depth, with late layers increasingly involved in forming the output~\citep{whatsinimg,knowledgeevolve,arbitration}. Our layerwise and range-averaged analyses reveal a clear boundary rather than a gradual progression. After this boundary, image-induced semantic updates become tightly linked to the corresponding answer revision.

\subsection{Causal Residual Interchange}
\label{sec:causal-interchange}
The clear late-coupling boundary motivates a causal test of the corresponding states in \qwenthreebshort{}. We intervene at one layer at a time, replacing only the normal run's final prompt-token state at that layer with its counterfactual counterpart. Using the output margin $t_p$ defined above, we compute the induced logit-margin update $\Delta t$ as the intervened margin minus the original normal-run margin. Positive values indicate movement toward GT. More details appear in Appendix~\ref{app:activation-interchange-details}.

\begin{figure}[h]
    \centering
    \includegraphics[width=0.95\linewidth,trim=0 10bp 0 10bp,clip]{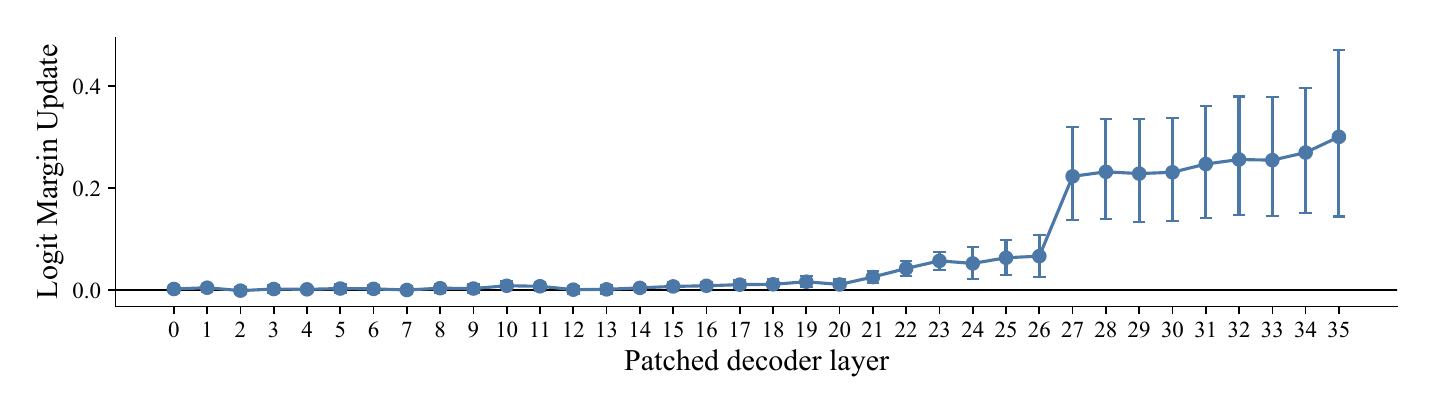}
    \caption{Causal residual interchange in \qwenthreeb{}. Each point is the mean logit-margin update $\Delta t$ after replacing the normal run's final prompt-token state with its counterfactual counterpart at a given layer. Positive values indicate movement toward GT; bars are 95\% confidence intervals.}
    \label{fig:activation-interchange}
\end{figure}

As \Figref{fig:activation-interchange} shows, residual interchange in \qwenthreebshort{} has little effect through the middle decoder and begins to shift the output margin around layers 21--23. Mean $\Delta t$ then more than triples at layer 27, rising from $0.067$ to $0.223$, and remains large through the final layers, reaching $0.301$ at the output. Range-level interventions reproduce this separation: layers 25--31 and 32--35 yield $\Delta t=0.247$ and $0.301$, respectively, while layers 4--10 have negligible influence. Replacing the normal run's late states with their counterfactual counterparts therefore causally shifts the answer margin toward GT.

\subsection{Module-Level Localization}
The abrupt jump at layer 27 motivates a finer localization within the layer. A Transformer layer contains self-attention and MLP submodules, each combined with the residual stream through a residual update~\citep{vaswani2017attention}. In the pre-norm Qwen architecture, each submodule receives an RMS-normalized residual state. We therefore repeat residual interchange at three boundaries in layers 26, 27, and 28: the layer input, the post-attention state, and the layer output. For all 197 Q1 pairs, we replace the normal run's final prompt-token state at each boundary with its counterfactual counterpart and allow the remaining computation to proceed from the replaced state. We compute the same logit-margin update $\Delta t$ as in Section~\ref{sec:causal-interchange}; paired differences in $\Delta t$ between adjacent boundaries measure the gain associated with the attention and MLP steps.

Table~\ref{tab:block27-boundaries} localizes the jump to the attention step of layer 27. Mean $\Delta t$ jumps sharply from $0.067$ at the layer input to $0.227$ after attention, then remains nearly unchanged at the layer output ($0.223$). The corresponding attention gain is $+0.160$, whereas the MLP gain is $-0.003$. The neighboring layers show no comparable attention gain, and only the layer-27 attention gain survives correction over the nine layer-by-step tests. The sharp causal transition in \qwenthreebshort{} therefore occurs primarily during attention in layer 27. Appendix~\ref{app:block27-case-study} provides the technical details.

\begin{table}[h]
    \centering
    \captionsetup{position=top}
    \caption{Module-level residual interchange in layers 26--28 of \qwenthreeb{}, evaluated on all 197 Q1 pairs. At each boundary, the normal run's final prompt-token state is replaced with its counterfactual counterpart. Values are mean logit-margin updates $\Delta t$, with positive values indicating movement toward GT. Gains are paired differences in $\Delta t$ between adjacent boundaries. Only the layer-27 attention gain, shown in bold, survives correction over the nine layer-by-step tests.}
    \label{tab:block27-boundaries}
    \footnotesize
    \setlength{\tabcolsep}{6pt}
    \renewcommand{\arraystretch}{1.3}
    \begin{tabular}{c | c c c | c c}
        \hline
        \textbf{Layer} & \textbf{Input} & \textbf{After attn.} & \textbf{Output} & \textbf{Attn. gain} (95\% CI) & \textbf{MLP gain} (95\% CI) \\
        \hline
        26 & 0.064 & 0.073 & 0.067 & $+0.009$ [$-0.007$, 0.026] & $-0.006$ [$-0.014$, 0.002] \\
        \textbf{27} & \textbf{0.067} & \textbf{0.227} & \textbf{0.223} & $+0.160$ [0.096, 0.233] & $-0.003$ [$-0.011$, 0.004] \\
        28 & 0.223 & 0.231 & 0.232 & $+0.008$ [$-0.015$, 0.032] & $+0.001$ [$-0.007$, 0.009] \\
        \hline
    \end{tabular}
\end{table}

\subsection{Cross-Architecture Generalization}
We test whether the observational internal--output coupling pattern found in \qwenthreebshort{} generalizes to \qwensevenbshort{} (28 layers) and \llavashort{} (32 layers). For each model, we train model-specific probes and repeat the paired regression, aligning layers by normalized decoder depth. Because the slope $\beta$ depends on each model's output scale, we compare models using the signed square root of partial $R^2$. This analysis tests observational coupling in all three models; causal residual interchange and module-level localization were evaluated in \qwenthreebshort{}.

\begin{figure}[h]
    \centering
    \includegraphics[width=0.90\linewidth,trim=0 10bp 0 10bp,clip]{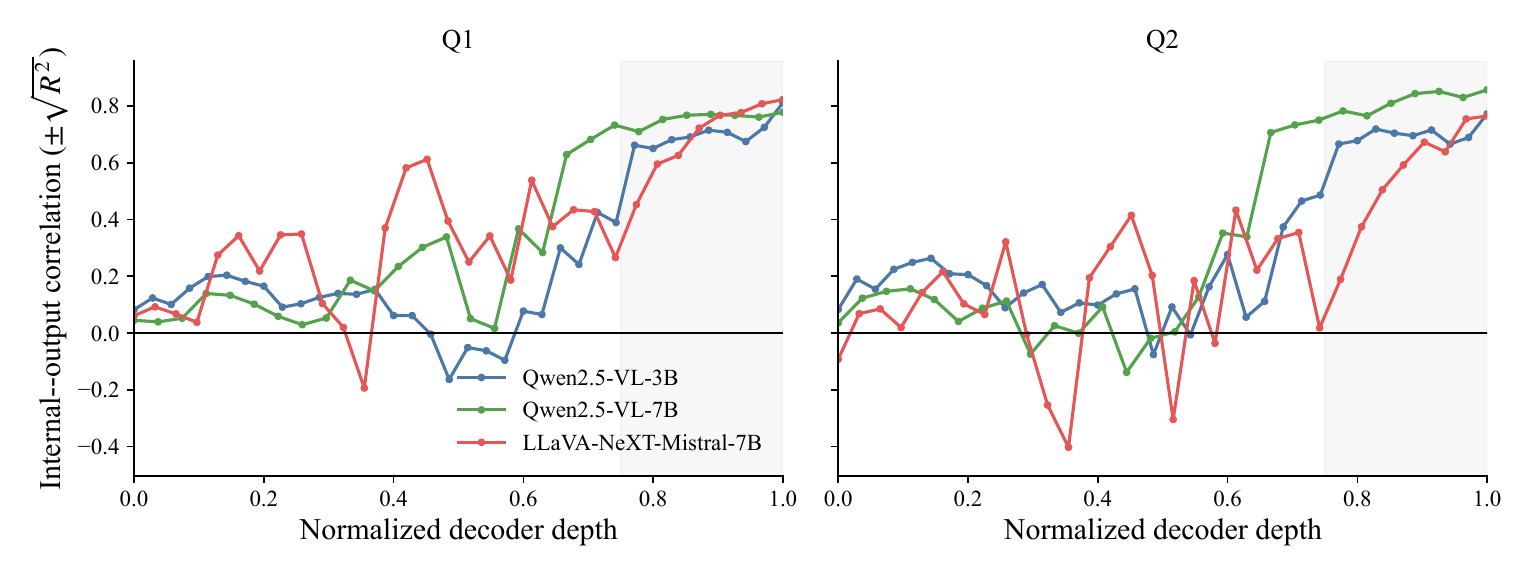}
    \caption{Cross-architecture internal--output update coupling for \qwenthreebshort{}, \qwensevenbshort{}, and \llavashort{}. The left and right panels show Q1 (dataset wording) and Q2 (matched command paraphrase), respectively, with layers aligned by normalized decoder depth. Each curve reports the signed square root of partial $R^2$, which is a correlation-like measure of the direction and strength of internal-output coupling. Shading marks the final quarter of the decoder.}
    \label{fig:cross-architecture}
\end{figure}

\Figref{fig:cross-architecture} shows weak and variable coupling across early and middle layers, followed by a sustained positive late band in all three models and under both question wordings. At the final layer, the Q1 values reach approximately $+0.81$, $+0.78$, and $+0.82$ for \qwenthreebshort{}, \qwensevenbshort{}, and \llavashort{}, respectively. The sustained positive late-layer band begins at normalized decoder depths $0.71$, $0.59$, and $0.77$, respectively. The band therefore begins at different normalized depths across architectures, while the late concentration of positive coupling is consistent across models and prompts. Model-specific training, tokenization, and statistical details appear in Appendix~\ref{app:cross-architecture-details}.

Together, these analyses separate the availability of answer-related information from its influence on prediction: the relevant answer may be decodable from intermediate states, while image-induced semantic updates remain weakly or variably coupled to output revision until late in the decoder. This pattern recurs across all three architectures as a sustained positive late-layer band, although the band begins at different normalized depths. In \qwenthreebshort{}, residual interchange further reveals a sharp transition in causal influence, with late states shifting the answer margin toward the visual ground truth and most of the increase occurring during the attention step of layer 27.

\FloatBarrier
\section{Visual Insensitivity Under Controlled Image Interventions}
\label{sec:behavioral}

Answer-related visual information may be represented within the decoder yet influence prediction only late, raising a behavioral question: can a model receive benchmark credit while remaining insensitive to the visual detail relevant to its answer? We test this across twelve benchmarks and three models by progressively removing image detail and measuring the fraction of answers that remain unchanged throughout substantial detail loss, together with the accuracy retained by this invariant subset. Sustained invariance helps determine which visual distinctions benchmark success actually requires, although an unchanged answer can remain justified when sufficient evidence survives.

\subsection{Progressive-Decay Evaluation}
\begin{figure}[t]
    \centering
    \begin{tcolorbox}[enhanced,arc=4pt,boxrule=0.6pt,top=3pt,bottom=3pt,left=3pt,right=3pt,colframe=borderblue,colback=white,hbox]
        \includegraphics[width=0.98\linewidth,trim=0 12bp 0 10bp,clip]{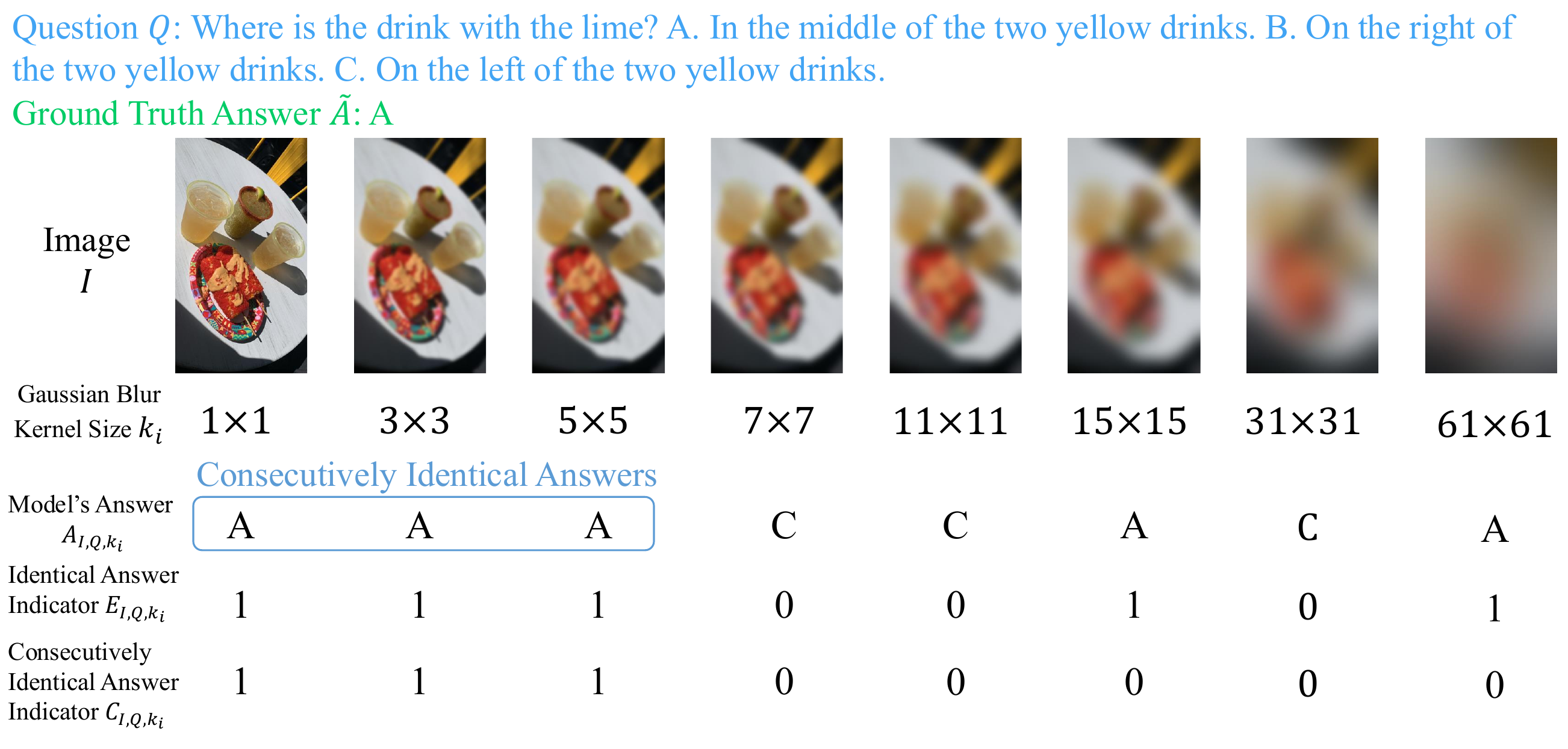}
    \end{tcolorbox}
    \caption{Progressive visual decay and consecutive agreement. A return to the original answer at a later blur level is excluded after the agreement chain first breaks.}
    \label{fig:blurs}
\end{figure}

We consider twelve benchmarks: RePOPE~\citep{repope}, HR-Bench~\citep{hrbench}, both HallusionBench partitions~\citep{hallusionbench}, AI2D~\citep{ai2d}, MMMU~\citep{mmmu}, V*Bench~\citep{vstar}, VLMBias~\citep{vlmbias}, RealworldQA~\citep{realworldqa}, BLINK~\citep{blink}, MMBench~\citep{mmbench}, SEED-Bench~\citep{seedbench}, and MMStar~\citep{mmstar}. For each image $I$ and question $Q$, we progressively apply Gaussian blur with kernels $K=(\kersizes)$. At the strongest blur setting, $\ksz{61}$, object outlines are completely indistinguishable, as illustrated in \Figref{fig:blurs}. Greedy decoding yields answer $A_{I,Q,k_i}$ at each level, with $k_1$ denoting the unblurred baseline. We define
\[
E_{I,Q,k_i}=\mathbb{1}[A_{I,Q,k_i}=A_{I,Q,k_1}],\qquad
C_{I,Q,k_i}=\bigwedge_{j=1}^{i}E_{I,Q,k_j}.
\]
$E_{I,Q,k_i}$ indicates whether the answer at the current blur level equals the original-image answer at $k_1$. As the logical AND operator $\bigwedge$ requires, $C_{I,Q,k_i}=1$ only when this equality holds at every level through $k_i$; once broken, the chain cannot be restored by a later return (\Figref{fig:blurs}). Under an illustrative independent uniform-guessing baseline for a binary question, the chance of matching the original answer falls from $1/2$ for one degraded image to $(1/2)^7$ across all seven degraded levels, reducing accidental matches in constrained answer spaces. We report dataset means $\dseqprob$ and $\dsconsprob$, overall accuracy $\dsbluracc$, and invariant-subset accuracy $\dsconsacc$; full definitions and settings appear in Appendix~\ref{app:behavioral-details}.

The two accuracy metrics answer different evaluation questions. At each blur level $k_i$, overall accuracy $\dsbluracc$ is computed over the full evaluation set with every image blurred to $k_i$, whereas invariant-subset accuracy $\dsconsacc$ is computed only over samples whose predictions have matched the original-image answer through levels $k_1,k_2,\ldots,k_i$. Because this subset shrinks with degradation, its accuracy can exceed the unblurred full-dataset accuracy. Such an increase reflects selection of invariant responses, not improved robustness of a fixed population. A high $\dsconsprob$ establishes persistence along the blur sequence, and a high $\dsconsacc$ establishes that those persistent responses receive benchmark credit. We therefore report invariant-subset accuracy together with the fraction of predictions that remain invariant.

\subsection{Experimental Results}
\begin{figure}[ht]
    \centering
    \includegraphics[width=\linewidth,trim=0 24bp 0 20bp,clip]{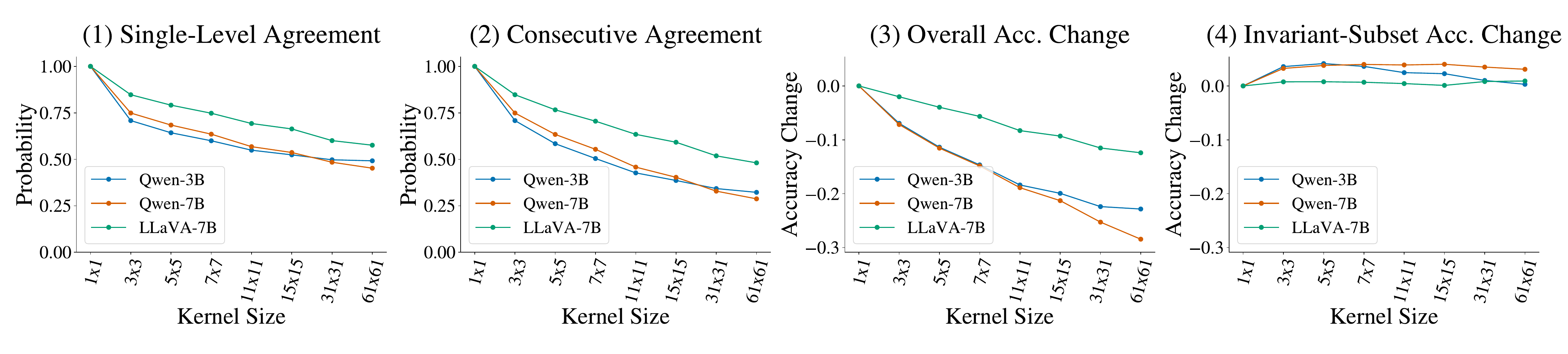}
    \caption{Dataset-averaged trajectories under increasing Gaussian blur for \qwenthreebshort{}, \qwensevenbshort{}, and \llavashort{}. Panel 1 shows single-level agreement: the answer at the current blur level $k_i$ equals the answer on the unblurred image $k_1$. Panel 2 shows consecutive agreement: the answer equals the unblurred answer at every blur level from $k_1$ through $k_i$. Panels 3 and 4 show overall accuracy change $\dsbluracc-\bar{\alpha}_{\ds,k_1}$ and invariant-subset accuracy change $\dsconsacc-\bar{\gamma}_{\ds,k_1}$, respectively. Because raw accuracies differ across benchmarks, each dataset's own $k_1$ accuracy is subtracted before averaging.}
    \label{fig:model-comparison}
\end{figure}

We evaluate progressive decay across the twelve benchmarks and three models introduced above. \Figref{fig:model-comparison} reports trajectories averaged across datasets, while Appendix \Figref{fig:app-dataset-results} provides the individual trajectories for each benchmark. Appendix~\ref{app:visual-decay-controls} reports complementary visual-decay controls, including grayscale, nested spatial masking, and text-only prompts, across the same three models and twelve benchmarks.

\paragraph{Many predictions persist under severe blur.}
As \Figref{fig:model-comparison} shows, substantial fractions of predictions remain unchanged throughout the blur sequence. At the strongest blur, dataset-averaged consecutive agreement remains $32.2\%$ for \qwenthreebshort{}, $28.7\%$ for \qwensevenbshort{}, and $48.1\%$ for \llavashort{}. Per-benchmark trajectories show particularly high persistence on RePOPE and substantial persistence on MMMU (\Figref{fig:ds-repope}; \Figref{fig:ds-appendix-mmmu}). Even when object outlines are indistinguishable, the models do not uniformly abstain or report that the image is unreadable. Many predictions instead retain the same scorable answer through all seven degraded levels, showing that the models do not fully use the visual detail removed by blur on a substantial subset of questions.

\paragraph{Consecutive agreement excludes trajectories that later return to the original answer.}
At the strongest blur, single-level and consecutive agreement are respectively $49.2\%$ and $32.2\%$ for \qwenthreebshort{}, $45.2\%$ and $28.7\%$ for \qwensevenbshort{}, and $57.6\%$ and $48.1\%$ for \llavashort{} (\Figref{fig:model-comparison}). The gaps represent answers that changed at an earlier blur level and later returned to the original answer, which endpoint-only matching would count as unchanged. By excluding these interrupted returns, consecutive agreement provides a more reliable conservative lower bound on detectable visual insensitivity.

\paragraph{Unchanged predictions retain high benchmark accuracy.}
On RePOPE, overall accuracy falls from $92.7\%$ to $56.8\%$ under the strongest blur, while invariant-subset accuracy remains at $93.3\%$ (\Figref{fig:ds-repope}). The dataset average shows the same separation. From the unblurred images to the strongest blur, overall accuracy decreases by $22.8$, $28.4$, and $12.4$ percentage points for \qwenthreebshort{}, \qwensevenbshort{}, and \llavashort{}, respectively, whereas invariant-subset accuracy changes by $+0.3$, $+3.1$, and $+0.9$ percentage points (\Figref{fig:model-comparison}). Thus, benchmark credit remains high among predictions that do not respond to the removed detail even as full-dataset performance declines. Together, these results show why accuracy alone does not establish use of the visual details affected by degradation.

\paragraph{Visually dependent questions lose invariant-subset accuracy faster.}
HallusionBench's Visual Dependent split loses invariant-subset accuracy faster than its Visual Supplement split for both Qwen models (\Figref{fig:ds-hallusionbench-vd}; \Figref{fig:app-ds-hallusionbench-vs}). This ordering is consistent with the greater importance of image content to correctness in the Visual Dependent split: unchanged predictions receive less credit as relevant evidence is degraded. The contrast supports the evaluation's sensitivity to differences in benchmark visual dependence.

\paragraph{Model size does not consistently predict invariance.}
In the dataset average (\Figref{fig:model-comparison}), the strongest blur reduces \llavashort{}'s accuracy by $12.4$ percentage points, compared with $22.8$ and $28.4$ percentage points for \qwenthreebshort{} and \qwensevenbshort{}, respectively; \llavashort{} also retains more original answers. Yet the Qwen 3B and 7B models exchange order across datasets, so larger size does not consistently reduce invariance. Lower degradation and higher consecutive agreement may reflect more severe visual insensitivity rather than stronger use of visual evidence.

\subsection{Qualitative Analysis}
\begin{figure}[h]
    \centering
    \includegraphics[width=0.8\linewidth,trim=0 6bp 0 8bp,clip]{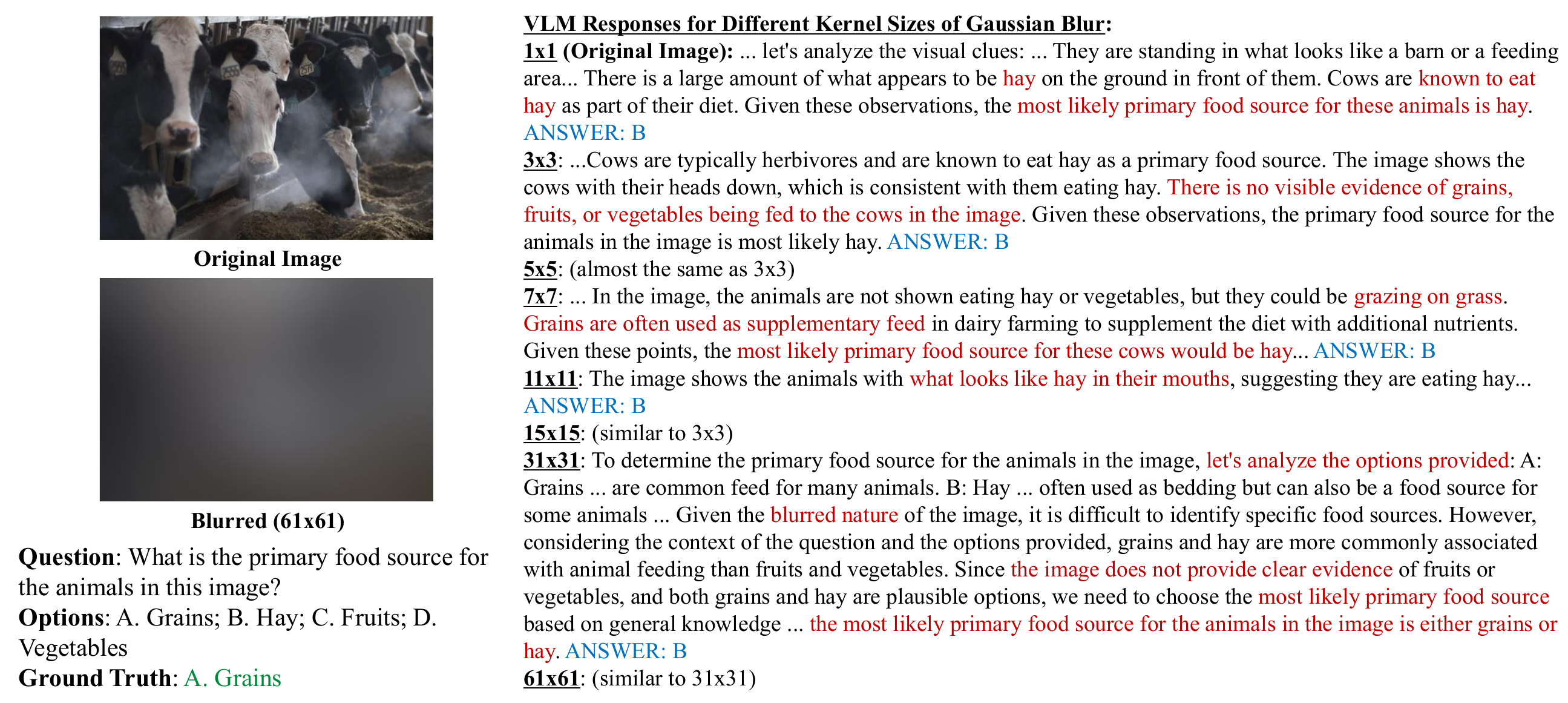}
    \caption{An invariant error from \qwensevenb{} in the MMStar~\citep{mmstar} dataset. The prediction remains \emph{hay} by general knowledge, despite the label \emph{grains}.}
    \label{fig:langprior-example-main}
\end{figure}

\Figref{fig:langprior-example-main} illustrates an invariant error from \qwensevenbshort{}. Regarding the cows' food, the model retains \emph{hay} as blur increases and shifts from visual claims to the familiar association between cows and hay, although the label is \emph{grains}. Appendix~\ref{app:behavioral-details} provides a complementary correct invariant response.

\section{Discussion and Conclusion}
\label{main-text-end}

\paragraph{Visual evidence availability and use.}
Many training distributions may allow a model to predict the answer from the question, coarse scene information, and familiar associations. When these signals are usually sufficient, there is little pressure to inspect the local details that determine the answer or preserve their influence through decoding. Our results show that answer-relevant detail can remain decodable while exerting limited influence on answer revision. This influence changes sharply across a clear late-layer boundary, including the abrupt attention-mediated jump at layer 27 in \qwenthreebshort{}, while coupling becomes strong only in late layers across models. A central direction for future work is to identify what determines this boundary and whether training can make detailed visual evidence influence predictions earlier and remain influential throughout decoding.

\paragraph{Implications for training.}
A training set remains vulnerable to this shortcut when the question and broad visual content reliably predict the target. Scaling similar data will continue to reward answers based on these coarse signals. Effective training distributions should make the target difficult to infer without the decisive visual detail, reduce the reward for relying on broad scene information, and maintain the influence of that detail as the answer is formed.

\paragraph{Implications for evaluation.}
A strong visual evaluation should ensure that the answer genuinely depends on the tested detail and cannot be reliably inferred from the text and broad scene content. Model behavior should change when the decisive evidence changes, while plausible guesses based on familiar associations should receive no advantage. Scoring should distinguish correct use of the relevant evidence from coincidental correctness, so that high performance reflects sensitivity to the visual information the task is intended to test.

Taken together, our findings underscore the importance of task-relevant visual evidence in shaping a model's output. Broad scene understanding provides useful context, while reliable vision-language reasoning also depends on the continued influence of the specific visual details that determine the appropriate response. Ultimately, an ideal VLM should accurately identify and use the visual details that matter for the task.

\subsection*{AI use statement}
We used generative AI tools only to assist with writing. Specifically, they were
used to edit the manuscript for readability, to tighten phrasing, and to help
refine the wording of the abstract, introduction, and related-work sections. All
suggestions were reviewed and revised by the authors, and every scientific claim
was checked against our experiments and the cited sources. We did not use
generative AI tools to generate data, develop the methodology, design or
implement experiments, produce proofs, or interpret results; these tasks are
either not applicable or were carried out entirely by the authors. We take
responsibility for the final content of this work, including all text, claims,
and artifacts.

\subsection*{Ethics statement}

This work evaluates publicly available vision-language models on existing
benchmarks, including released counterfactual image pairs. We do not collect
new data or infer information about individuals. Our interventions are limited
to controlled image degradations and state replacements for analysis. The goal
is to improve whether models use task-relevant visual evidence. Any deployment
of more reliable vision-language systems should nevertheless be evaluated for
application-specific risks, including errors that may affect people.

\subsection*{Reproducibility statement}

Sections~\ref{sec:internal} and~\ref{sec:behavioral} specify the paired
counterfactual analysis, residual interchange procedure, progressive-decay
evaluation, and consecutive-agreement metric. The appendix reports the
fine-tuning and layer-replacement setup, probe architecture and data splits,
paired-regression details, intervention definitions, model-specific replication
settings, visual-decay controls, and per-benchmark trajectories. We use
publicly available model checkpoints and benchmark data; prompts, sample counts,
random seeds, hyperparameters, and statistical procedures are specified in the
corresponding sections and appendices.

\bibliography{iclr2027_conference}

@inproceedings{vaswani2017attention,
  title={Attention Is All You Need},
  author={Vaswani, Ashish and Shazeer, Noam and Parmar, Niki and Uszkoreit, Jakob and Jones, Llion and Gomez, Aidan N. and Kaiser, Lukasz and Polosukhin, Illia},
  booktitle={Advances in Neural Information Processing Systems},
  volume={30},
  pages={5998--6008},
  year={2017},
  url={https://papers.nips.cc/paper/7181-attention-is-all-you-need}
}

@misc{qwen25,
      title={Qwen2.5-VL Technical Report}, 
      author={Shuai Bai and Keqin Chen and Xuejing Liu and Jialin Wang and Wenbin Ge and Sibo Song and Kai Dang and Peng Wang and Shijie Wang and Jun Tang and Humen Zhong and Yuanzhi Zhu and Mingkun Yang and Zhaohai Li and Jianqiang Wan and Pengfei Wang and Wei Ding and Zheren Fu and Yiheng Xu and Jiabo Ye and Xi Zhang and Tianbao Xie and Zesen Cheng and Hang Zhang and Zhibo Yang and Haiyang Xu and Junyang Lin},
      year={2025},
      eprint={2502.13923},
      archivePrefix={arXiv},
      primaryClass={cs.CV},
      url={https://arxiv.org/abs/2502.13923}, 
}

@misc{vlmbias,
      title={Vision Language Models are Biased}, 
      author={An Vo and Khai-Nguyen Nguyen and Mohammad Reza Taesiri and Vy Tuong Dang and Anh Totti Nguyen and Daeyoung Kim},
      year={2025},
      eprint={2505.23941},
      archivePrefix={arXiv},
      primaryClass={cs.LG},
      url={https://arxiv.org/abs/2505.23941}, 
}

@inproceedings{whatsinimg,
  title={What's in the Image? A Deep-Dive into the Vision of Vision Language Models},
  author={Kaduri, Omri and Bagon, Shai and Dekel, Tali},
  booktitle={Proceedings of the Computer Vision and Pattern Recognition Conference},
  pages={14549--14558},
  year={2025}
}

@article{deepseekmath,
  title={Deepseekmath: Pushing the limits of mathematical reasoning in open language models},
  author={Shao, Zhihong and Wang, Peiyi and Zhu, Qihao and Xu, Runxin and Song, Junxiao and Bi, Xiao and Zhang, Haowei and Zhang, Mingchuan and Li, YK and Wu, Yang and others},
  journal={arXiv preprint arXiv:2402.03300},
  year={2024}
}

@article{lora,
  title={Lora: Low-rank adaptation of large language models.},
  author={Hu, Edward J and Shen, Yelong and Wallis, Phillip and Allen-Zhu, Zeyuan and Li, Yuanzhi and Wang, Shean and Wang, Liang and Chen, Weizhu and others},
  journal={International Conference on Learning Representations},
  volume={1},
  number={2},
  pages={3},
  year={2022}
}

@inproceedings{knowledgeevolve,
  title={Towards understanding how knowledge evolves in large vision-language models},
  author={Wang, Sudong and Zhang, Yunjian and Zhu, Yao and Li, Jianing and Wang, Zizhe and Liu, Yanwei and Ji, Xiangyang},
  booktitle={Proceedings of the Computer Vision and Pattern Recognition Conference},
  pages={29858--29868},
  year={2025}
}

@misc{accelerate,
  title =        {Accelerate: Training and inference at scale made simple, efficient and adaptable.},
  author =       {Sylvain Gugger and Lysandre Debut and Thomas Wolf and Philipp Schmid and Zachary Mueller and Sourab Mangrulkar and Marc Sun and Benjamin Bossan},
  howpublished = {\url{https://github.com/huggingface/accelerate}},
  year =         {2022}
}

@misc{logitlens,
  title={Interpreting GPT: The Logit Lens},
  author={Nostalgebraist},
  year={2020},
  month={Aug},
  howpublished={\url{https://www.alignmentforum.org/posts/AcKRB8wDpdaN6v6ru/interpreting-gpt-the-logit-lens}},
}

@article{sae,
  title={K-sparse autoencoders},
  author={Makhzani, Alireza and Frey, Brendan},
  journal={arXiv preprint arXiv:1312.5663},
  year={2013}
}

@article{saellm,
  title={Sparse autoencoders find highly interpretable features in language models},
  author={Cunningham, Hoagy and Ewart, Aidan and Riggs, Logan and Huben, Robert and Sharkey, Lee},
  journal={arXiv preprint arXiv:2309.08600},
  year={2023}
}

@inproceedings{pixmo,
  title={Molmo and pixmo: Open weights and open data for state-of-the-art vision-language models},
  author={Deitke, Matt and Clark, Christopher and Lee, Sangho and Tripathi, Rohun and Yang, Yue and Park, Jae Sung and Salehi, Mohammadreza and Muennighoff, Niklas and Lo, Kyle and Soldaini, Luca and others},
  booktitle={Proceedings of the Computer Vision and Pattern Recognition Conference},
  pages={91--104},
  year={2025}
}

@inproceedings{llava,
  title={Improved baselines with visual instruction tuning},
  author={Liu, Haotian and Li, Chunyuan and Li, Yuheng and Lee, Yong Jae},
  booktitle={Proceedings of the IEEE/CVF conference on computer vision and pattern recognition},
  pages={26296--26306},
  year={2024}
}

@inproceedings{blink,
  title={BLINK: Multimodal Large Language Models Can See but Not Perceive},
  author={Fu, Xingyu and Hu, Yushi and Li, Bangzheng and Feng, Yu and Wang, Haoyu and Lin, Xudong and Roth, Dan and Smith, Noah A and Ma, Wei-Chiu and Krishna, Ranjay},
  booktitle={European Conference on Computer Vision},
  pages={148--166},
  year={2024}
}

@misc{realworldqa,
  author = {{X.AI}},
  title = {Grok-1.5 vision preview},
  year = {2024},
  howpublished = {\url{https://x.ai/blog/grok-1.5v}},
}

@article{repope,
  title={Repope: Impact of annotation errors on the pope benchmark},
  author={Neuhaus, Yannic and Hein, Matthias},
  journal={arXiv preprint arXiv:2504.15707},
  year={2025}
}

@inproceedings{vstar,
  title={V*: Guided visual search as a core mechanism in multimodal llms},
  author={Wu, Penghao and Xie, Saining},
  booktitle={Proceedings of the IEEE/CVF Conference on Computer Vision and Pattern Recognition},
  pages={13084--13094},
  year={2024}
}

@inproceedings{hallusionbench,
  title={Hallusionbench: an advanced diagnostic suite for entangled language hallucination and visual illusion in large vision-language models},
  author={Guan, Tianrui and Liu, Fuxiao and Wu, Xiyang and Xian, Ruiqi and Li, Zongxia and Liu, Xiaoyu and Wang, Xijun and Chen, Lichang and Huang, Furong and Yacoob, Yaser and others},
  booktitle={Proceedings of the IEEE/CVF conference on computer vision and pattern recognition},
  pages={14375--14385},
  year={2024}
}

@inproceedings{mmmu,
  title={Mmmu: A massive multi-discipline multimodal understanding and reasoning benchmark for expert agi},
  author={Yue, Xiang and Ni, Yuansheng and Zhang, Kai and Zheng, Tianyu and Liu, Ruoqi and Zhang, Ge and Stevens, Samuel and Jiang, Dongfu and Ren, Weiming and Sun, Yuxuan and others},
  booktitle={Proceedings of the IEEE/CVF conference on computer vision and pattern recognition},
  pages={9556--9567},
  year={2024}
}

@inproceedings{ai2d,
  title={A diagram is worth a dozen images},
  author={Kembhavi, Aniruddha and Salvato, Mike and Kolve, Eric and Seo, Minjoon and Hajishirzi, Hannaneh and Farhadi, Ali},
  booktitle={European conference on computer vision},
  pages={235--251},
  year={2016},
  organization={Springer}
}

@inproceedings{hrbench,
  title={Divide, conquer and combine: A training-free framework for high-resolution image perception in multimodal large language models},
  author={Wang, Wenbin and Ding, Liang and Zeng, Minyan and Zhou, Xiabin and Shen, Li and Luo, Yong and Yu, Wei and Tao, Dacheng},
  booktitle={Proceedings of the AAAI Conference on Artificial Intelligence},
  volume={39},
  pages={7907--7915},
  year={2025}
}

@article{mmstar,
  title={Are we on the right way for evaluating large vision-language models?},
  author={Chen, Lin and Li, Jinsong and Dong, Xiaoyi and Zhang, Pan and Zang, Yuhang and Chen, Zehui and Duan, Haodong and Wang, Jiaqi and Qiao, Yu and Lin, Dahua and others},
  journal={Advances in Neural Information Processing Systems},
  volume={37},
  pages={27056--27087},
  year={2024}
}

@inproceedings{mmbench,
  title={Mmbench: Is your multi-modal model an all-around player?},
  author={Liu, Yuan and Duan, Haodong and Zhang, Yuanhan and Li, Bo and Zhang, Songyang and Zhao, Wangbo and Yuan, Yike and Wang, Jiaqi and He, Conghui and Liu, Ziwei and others},
  booktitle={European conference on computer vision},
  pages={216--233},
  year={2024},
  organization={Springer}
}

@article{seedbench,
  title={Seed-bench: Benchmarking multimodal llms with generative comprehension},
  author={Li, Bohao and Wang, Rui and Wang, Guangzhi and Ge, Yuying and Ge, Yixiao and Shan, Ying},
  journal={arXiv preprint arXiv:2307.16125},
  year={2023}
}

@InProceedings{langprior2023,
  title = 	 {Revisiting the Role of Language Priors in Vision-Language Models},
  author =       {Lin, Zhiqiu and Chen, Xinyue and Pathak, Deepak and Zhang, Pengchuan and Ramanan, Deva},
  booktitle = 	 {Proceedings of the 41st International Conference on Machine Learning},
  pages = 	 {29914--29934},
  year = 	 {2024},
  editor = 	 {Salakhutdinov, Ruslan and Kolter, Zico and Heller, Katherine and Weller, Adrian and Oliver, Nuria and Scarlett, Jonathan and Berkenkamp, Felix},
  volume = 	 {235},
  series = 	 {Proceedings of Machine Learning Research},
  month = 	 {21--27 Jul},
  publisher =    {PMLR},
  url = 	 {https://proceedings.mlr.press/v235/lin24c.html}
}

@inproceedings{vlindbench,
  title={Vlind-bench: Measuring language priors in large vision-language models},
  author={Lee, Kang-il and Kim, Minbeom and Yoon, Seunghyun and Kim, Minsung and Lee, Dongryeol and Koh, Hyukhun and Jung, Kyomin},
  booktitle={Findings of the Association for Computational Linguistics: NAACL 2025},
  pages={4129--4144},
  year={2025}
}

@inproceedings{vilp,
  title={Probing Visual Language Priors in VLMs},
  author={Luo, Tiange and Cao, Ang and Lee, Gunhee and Johnson, Justin and Lee, Honglak},
  booktitle={International Conference on Machine Learning},
  pages={41120--41156},
  year={2025},
  organization={PMLR}
}

@inproceedings{pixelsvspriors,
  title={Pixels versus priors: Controlling knowledge priors in vision-language models through visual counterfacts},
  author={Golovanevsky, Michal and Rudman, William and Lepori, Michael A and Bar, Amir and Singh, Ritambhara and Eickhoff, Carsten},
  booktitle={Proceedings of the 2025 Conference on Empirical Methods in Natural Language Processing},
  pages={24848--24863},
  year={2025}
}

@inproceedings{blindfaithtext,
  title={Words or vision: Do vision-language models have blind faith in text?},
  author={Deng, Ailin and Cao, Tri and Chen, Zhirui and Hooi, Bryan},
  booktitle={Proceedings of the Computer Vision and Pattern Recognition Conference},
  pages={3867--3876},
  year={2025}
}

@inproceedings{vlmblind,
  title={Vision language models are blind},
  author={Rahmanzadehgervi, Pooyan and Bolton, Logan and Taesiri, Mohammad Reza and Nguyen, Anh Totti},
  booktitle={Proceedings of the Asian Conference on Computer Vision},
  pages={18--34},
  year={2024}
}

@misc{mirage,
      title={MIRAGE: The Illusion of Visual Understanding}, 
      author={Mohammad Asadi and Jack W. O'Sullivan and Fang Cao and Tahoura Nedaee and Kamyar Rajabalifardi and Fei-Fei Li and Ehsan Adeli and Euan Ashley},
      year={2026},
      eprint={2603.21687},
      archivePrefix={arXiv},
      primaryClass={cs.AI},
      url={https://arxiv.org/abs/2603.21687}, 
}

@article{tunedlens,
  title={Eliciting latent predictions from transformers with the tuned lens},
  author={Belrose, Nora and Ostrovsky, Igor and McKinney, Lev and Furman, Zach and Smith, Logan and Halawi, Danny and Biderman, Stella and Steinhardt, Jacob},
  journal={arXiv preprint arXiv:2303.08112},
  year={2023}
}

@article{saevlm,
  title={Sparse autoencoders learn monosemantic features in vision-language models},
  author={Pach, Mateusz and Karthik, Shyamgopal and Bouniot, Quentin and Belongie, Serge and Akata, Zeynep},
  journal={Advances in Neural Information Processing Systems},
  volume={38},
  pages={95706--95742},
  year={2026}
}

@article{hiddenplainsight,
  title={Hidden in plain sight: Vlms overlook their visual representations},
  author={Fu, Stephanie and Bonnen, Tyler and Guillory, Devin and Darrell, Trevor},
  journal={arXiv preprint arXiv:2506.08008},
  year={2025}
}

@article{arbitration,
  title={Arbitration Failure, Not Perceptual Blindness: How Vision-Language Models Resolve Visual-Linguistic Conflicts},
  author={Nooralahzadeh, Farhad and Rohanian, Omid and Zhang, Yi and F{\"u}rst, Jonathan and Stockinger, Kurt},
  journal={arXiv preprint arXiv:2604.09364},
  year={2026}
}

@inproceedings{interpreting,
  title={Interpreting and editing vision-language representations to mitigate hallucinations},
  author={Jiang, Nick and Kachinthaya, Anish and Petryk, Suzanne and Gandelsman, Yossi},
  booktitle={International Conference on Learning Representations},
  volume={2025},
  pages={63582--63605},
  year={2025}
}

@inproceedings{devils,
  title={Devils in middle layers of large vision-language models: Interpreting, detecting and mitigating object hallucinations via attention lens},
  author={Jiang, Zhangqi and Chen, Junkai and Zhu, Beier and Luo, Tingjin and Shen, Yankun and Yang, Xu},
  booktitle={Proceedings of the IEEE/CVF Conference on Computer Vision and Pattern Recognition},
  pages={25004--25014},
  year={2025}
}

@article{instructblip,
  title={Instructblip: Towards general-purpose vision-language models with instruction tuning},
  author={Dai, Wenliang and Li, Junnan and Li, Dongxu and Tiong, Anthony and Zhao, Junqi and Wang, Weisheng and Li, Boyang and Fung, Pascale N and Hoi, Steven},
  journal={Advances in neural information processing systems},
  volume={36},
  pages={49250--49267},
  year={2023}
}
\bibliographystyle{iclr2027_conference}

\newpage
\appendix

\section{GRPO Fine-Tuning and Layer Replacement}
\label{app:coarse-internal}

\begin{figure}[htb]
    \centering
    \begin{subfigure}[b]{0.30\linewidth}
        \includegraphics[width=\linewidth]{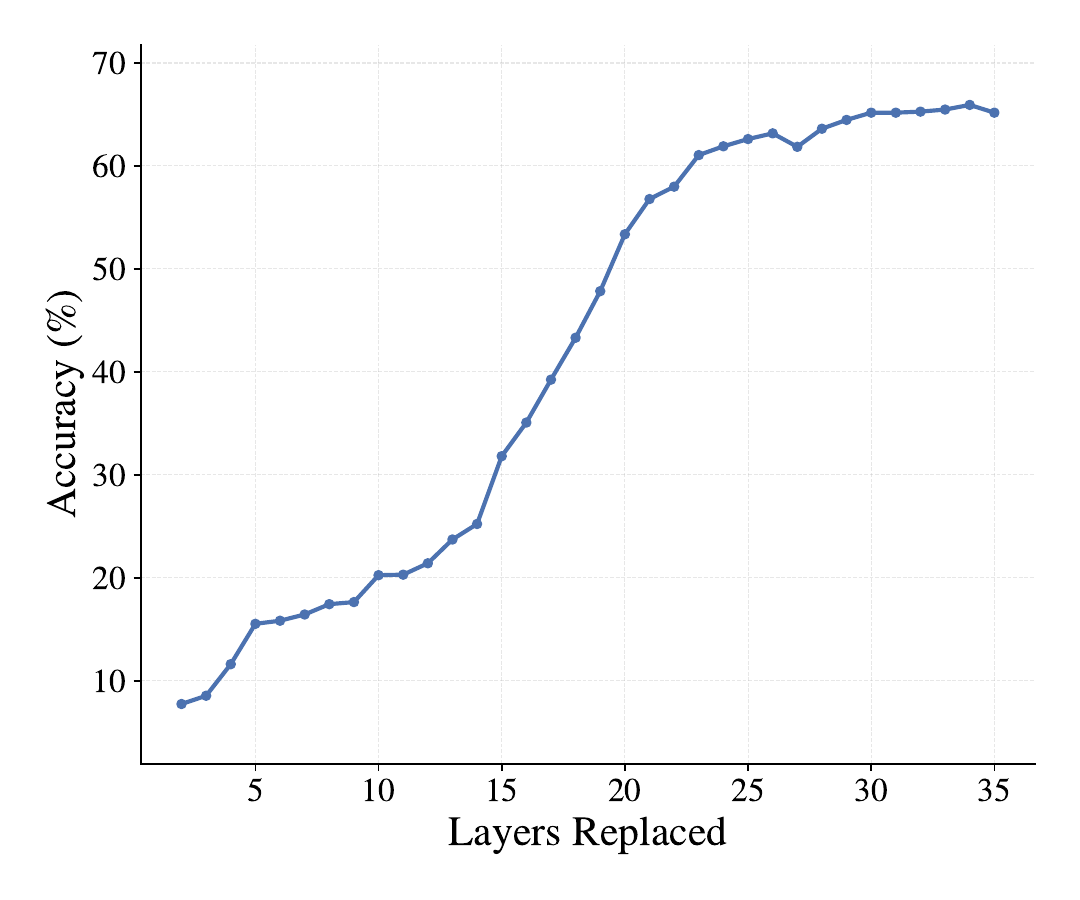}
        \caption{Accuracy versus replaced suffix.}
        \label{fig:replace-acc}
    \end{subfigure}
    \hfill
    \begin{subfigure}[b]{0.30\linewidth}
        \includegraphics[width=\linewidth]{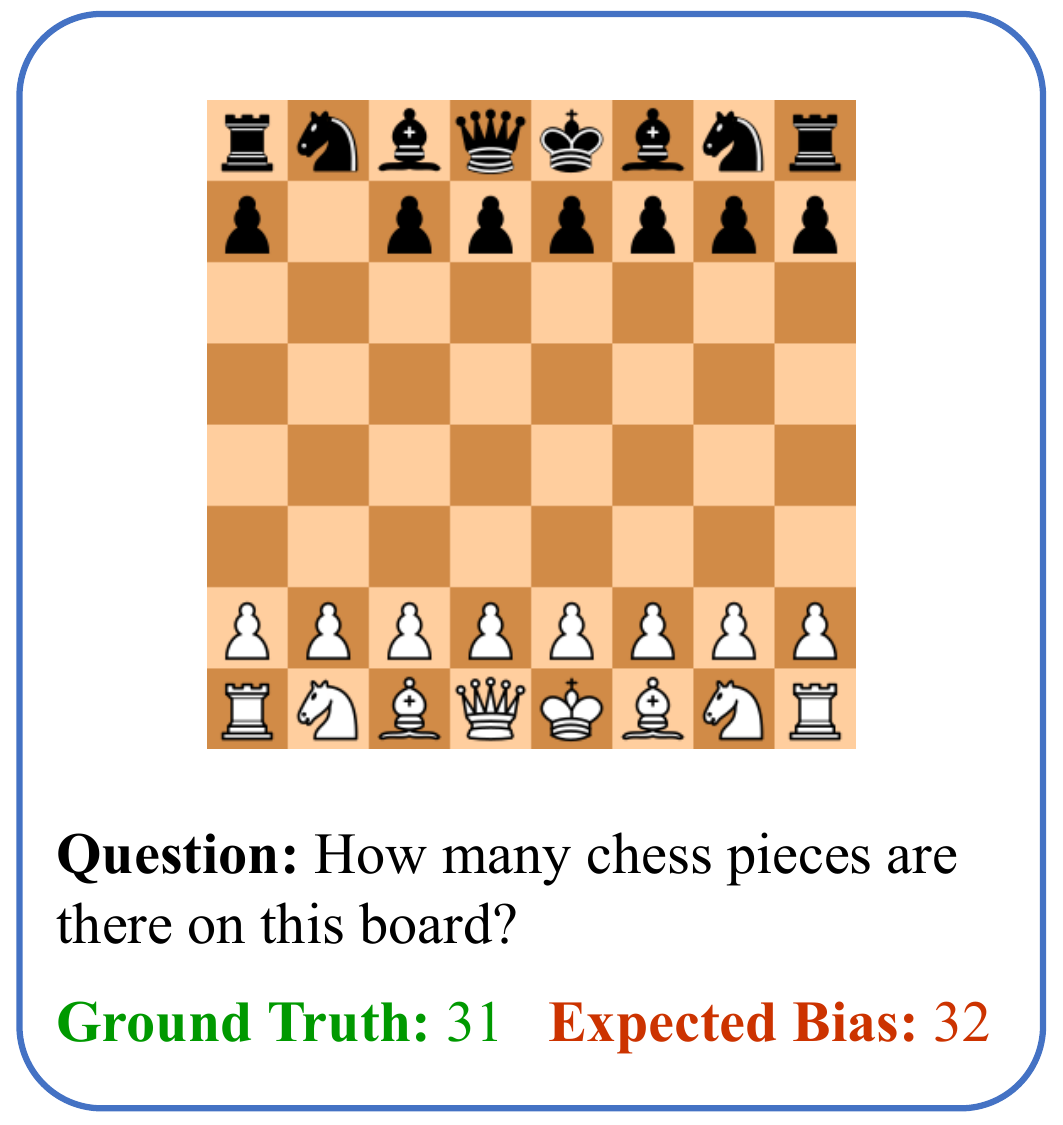}
        \caption{Corrected by five layers.}
        \label{fig:vlmbias-example-5ly}
    \end{subfigure}
    \hfill
    \begin{subfigure}[b]{0.30\linewidth}
        \includegraphics[width=\linewidth]{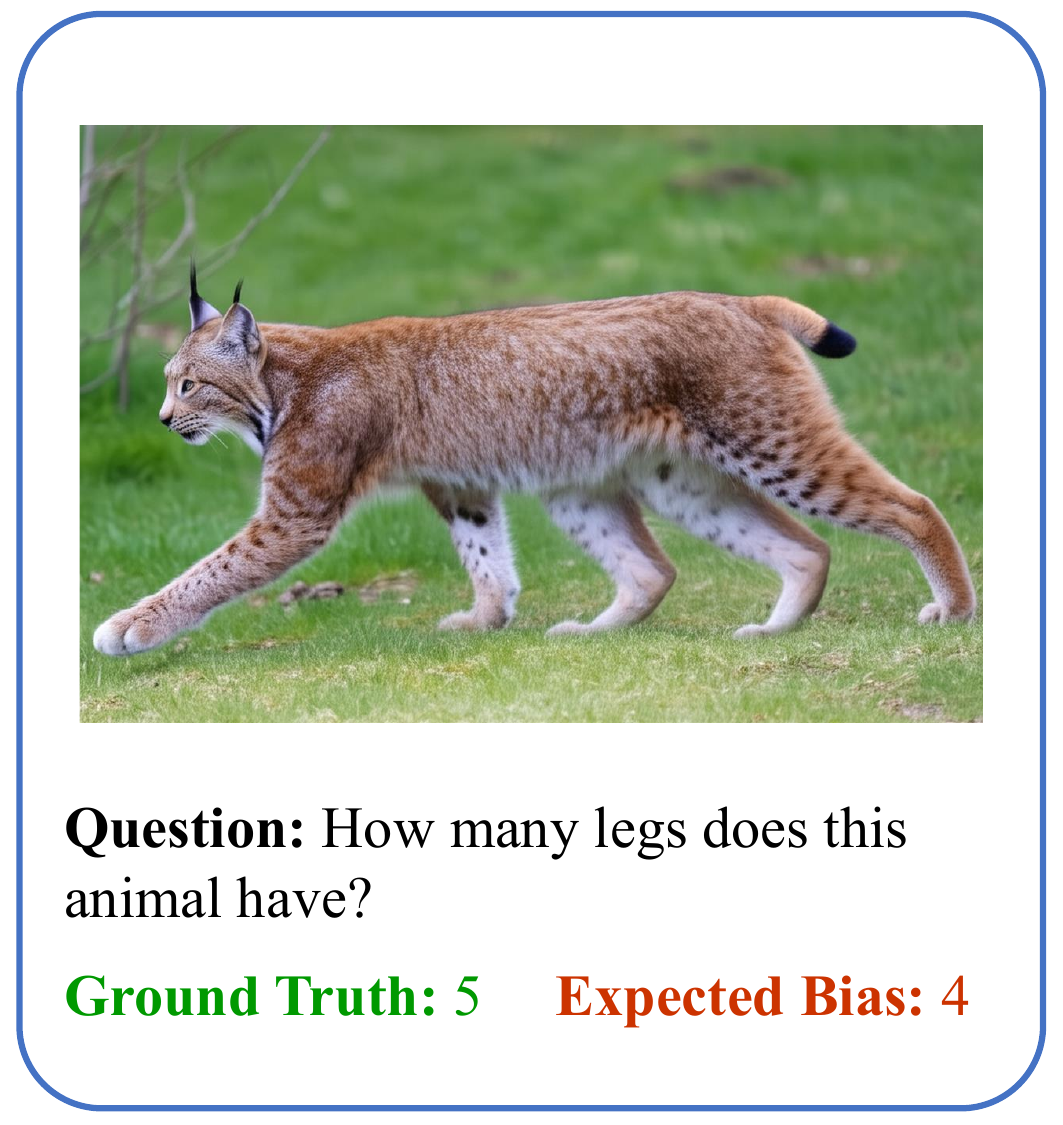}
        \caption{Corrected by twenty layers.}
        \label{fig:vlmbias-example-20ly}
    \end{subfigure}
    \caption{Coarse layer replacement and representative VLMBias examples. Additional examples appear in Appendix~\ref{app:vlmbias-more-examples}.}
    \label{fig:layer-replacement-analysis}
\end{figure}

We fine-tuned a debiased \qwenthreeb{} with GRPO and LoRA on VLMBias (Appendix~\ref{app:vlmbias-grpo}) and replaced suffixes of the baseline decoder with the corresponding fine-tuned layers. \Figref{fig:layer-replacement-analysis} shows that accuracy rises from 11.6\% to 65.2\% as more layers are replaced, with the largest increase when replacing the final 14--20 layers and a smaller increase for the final 3--5 layers; its two examples illustrate cases corrected by five and twenty layers. Replacing only the middle range improves accuracy to 24.3\%, suggesting that neither middle nor late layers alone explain the failure. Because hybrid networks can suffer representation mismatch, we use this as coarse causal localization rather than precise layer attribution.

\FloatBarrier
\section{Training Setup}
\label{app:vlmbias-grpo}

This section details the GRPO fine-tuning run whose layers are transplanted in Appendix~\ref{app:coarse-internal}. We use the Qwen2.5-VL-3B-Instruct~\citep{qwen25} model as our base vision-language model. The training is conducted using Group Relative Policy Optimization (GRPO)~\citep{deepseekmath}, a reinforcement learning algorithm that optimizes the model policy based on relative rewards within sampled groups. The experiments are executed on a cluster of 4 NVIDIA RTX 5880 Ada Generation GPUs using the Accelerate~\citep{accelerate} framework for distributed processing. We employ LoRA (Low-Rank Adaptation)~\citep{lora} for parameter-efficient fine-tuning, targeting all major projections in the language model backbone (including the QKV projection layers in the attention modules and the MLP layers) with a rank $r=32$ and $\alpha=32$.

The model is trained for $8$ epochs with a global batch size of 64 and a learning rate of $1 \times 10^{-5}$ following a cosine decay schedule. To ensure training stability, we include a warmup phase of 100 steps. During the GRPO rollout, we set the temperature to $1.0$ and top\_p to $0.9$, generating $G=8$ completions per prompt to calculate relative advantages.

Our reward function is a composite metric designed to enforce both structural adherence and factual accuracy:

\begin{itemize}
    \item Format Reward: Points are awarded for the inclusion of required markers (1. through 5.) and the specific curly brace delimiters.
    \item Accuracy Reward: A significant weight (75\% of the total potential reward) is assigned to the correctness of the final extracted answer. Correctness is determined by exact string matching or numerical equivalence, ensuring the model's reasoning culminates in an accurate conclusion.
\end{itemize}

For the training dataset, we identify matching normal-counterfactual image pairs in the VLMBias dataset~\citep{vlmbias}, e.g. a 2-legged ostrich image and a synthetic 3-legged counterpart. We include these images and the corresponding questions (e.g. How many legs does this animal have?) in the training dataset. To enhance the diversity of the training distribution and improve model robustness, we apply perceptually-conservative augmentations to each image. Specifically, we generate 30 augmented variations per original image using a stochastic pipeline of color jitter, ISO noise, and JPEG compression. This process results in a curated dataset of 9,300 samples, where the semantic integrity and ground-truth labels of the visual reasoning tasks remain preserved despite the introduced low-level image perturbations.

To induce structured chain-of-thought (CoT) reasoning, we modify the system prompt to require a five-part response:

\begin{enumerate}
    \item Object identification.
    \item Visual cue description.
    \item Evidence-to-question mapping.
    \item Visual reasoning derivation.
    \item Final answer enclosed in curly braces (e.g., \{answer\}).
\end{enumerate}

\FloatBarrier
\section{More Examples for Layer Replacement}
\label{app:vlmbias-more-examples}

\begin{figure}[hbt]
    \centering

    \begin{subfigure}[b]{0.4\linewidth}
        \centering
        \includegraphics[width=\linewidth]{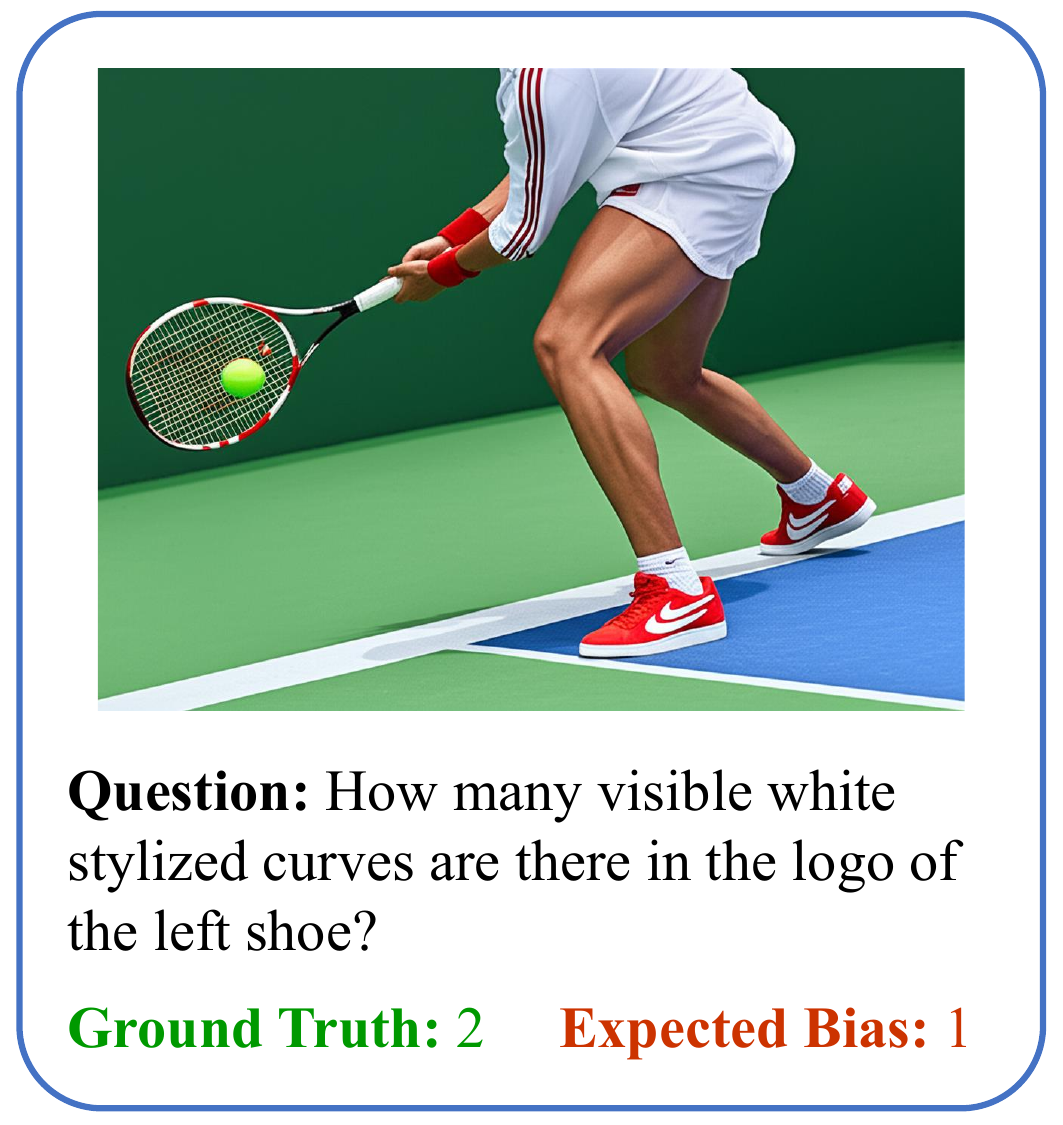}
    \end{subfigure}
    \hspace{0.1\linewidth}
    \begin{subfigure}[b]{0.4\linewidth}
        \centering
        \includegraphics[width=\linewidth]{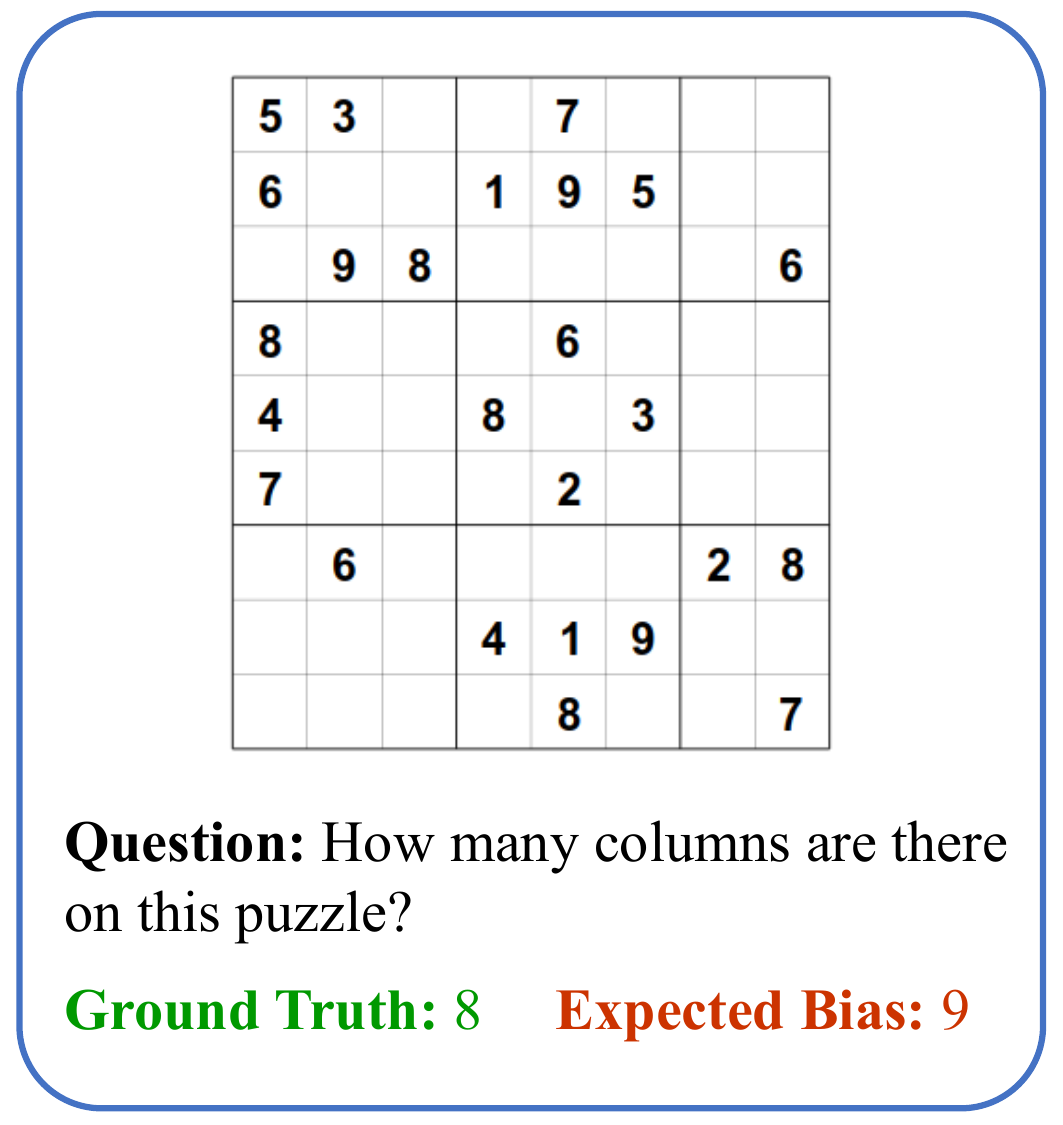}
    \end{subfigure}

    \caption{Questions related to the last 5 layers.}
    \label{fig:app-replace5ly}
\end{figure}

\begin{figure}[hbt]
    \centering

    \begin{subfigure}[b]{0.4\linewidth}
        \centering
        \includegraphics[width=\linewidth]{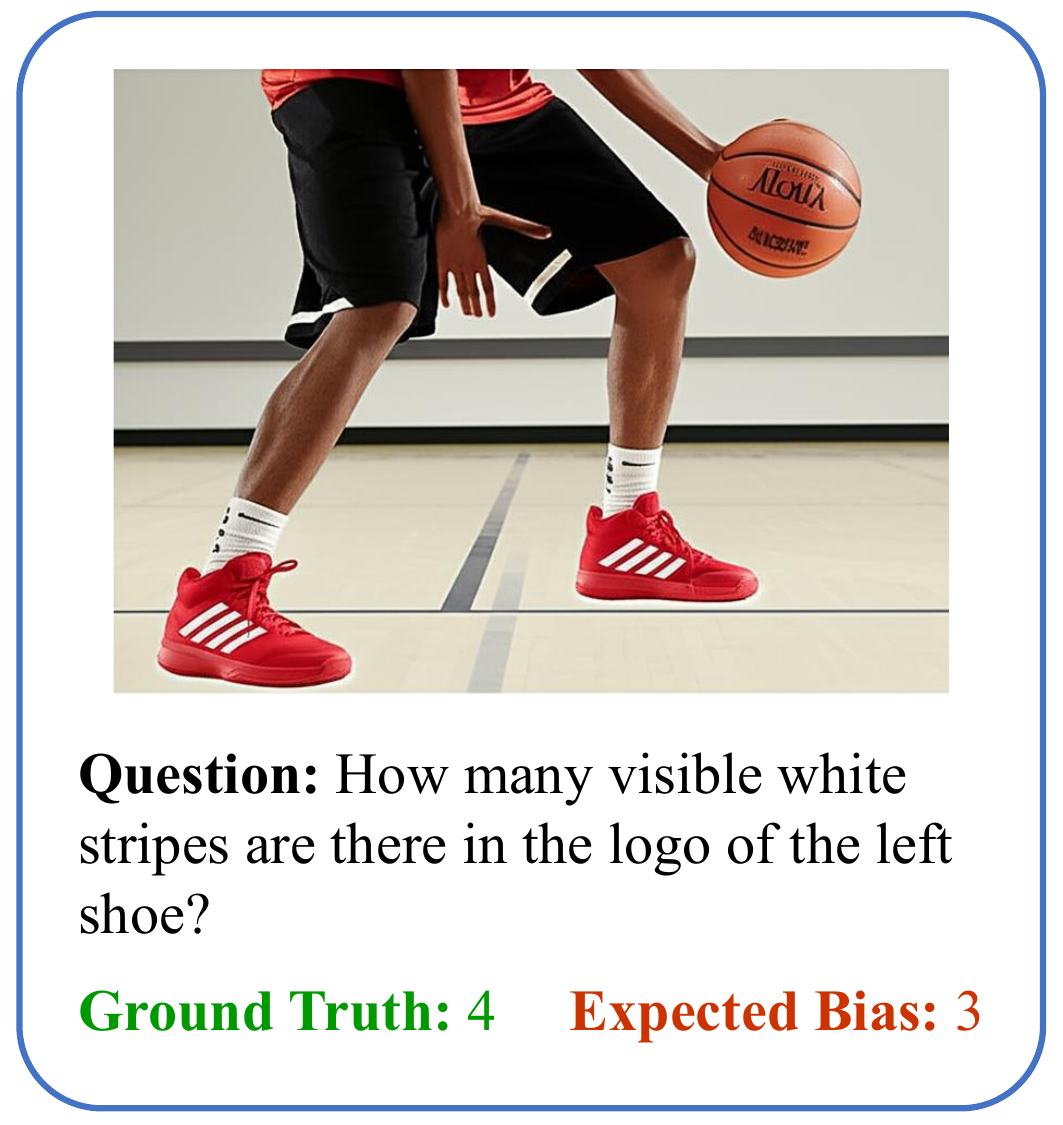}
    \end{subfigure}
    \hspace{0.1\linewidth}
    \begin{subfigure}[b]{0.4\linewidth}
        \centering
        \includegraphics[width=\linewidth]{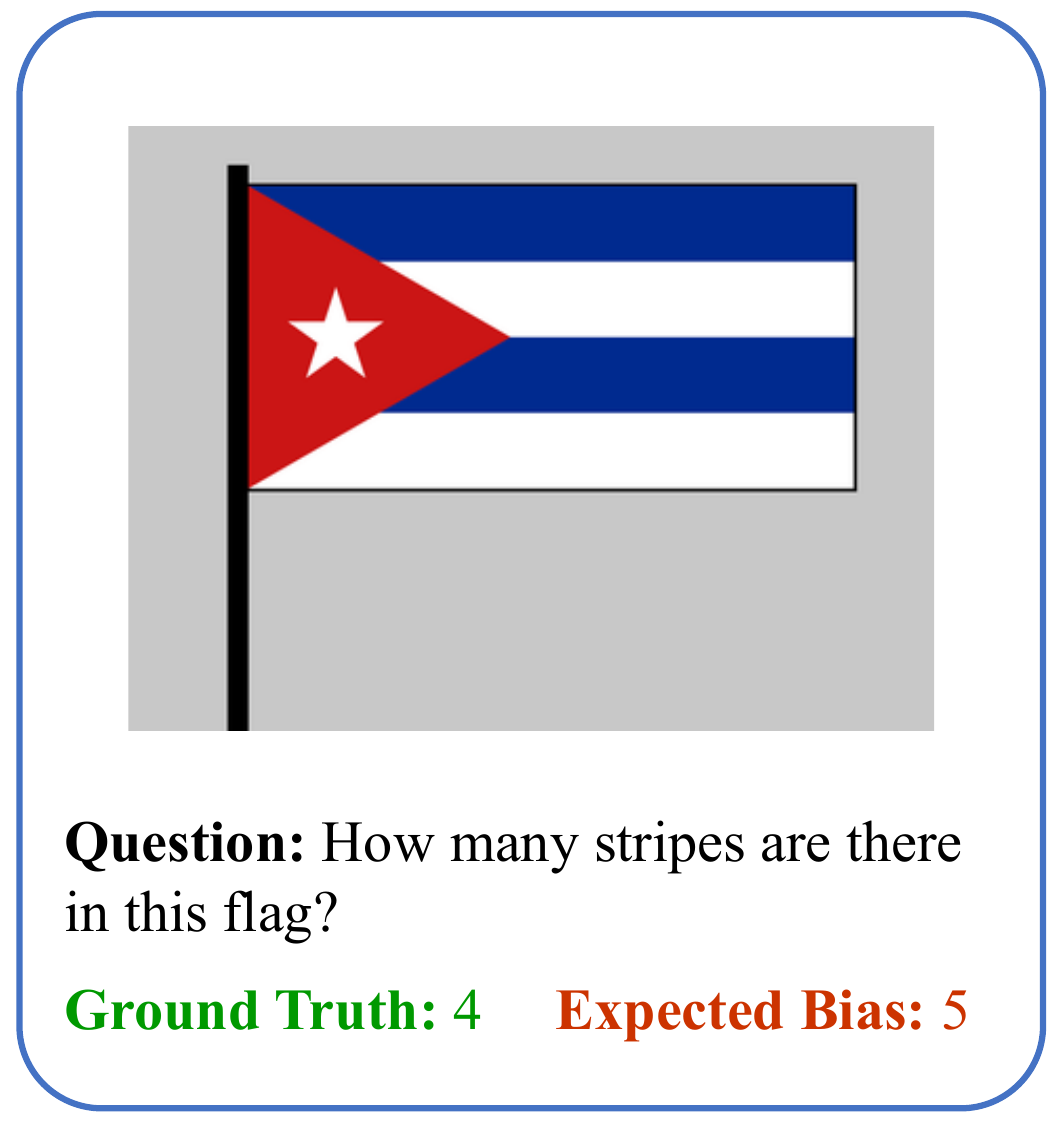}
    \end{subfigure}

    \caption{Questions related to the last 20 layers.}
    \label{fig:app-replace20ly}
\end{figure}

This section supplements the coarse layer-replacement study in Appendix~\ref{app:coarse-internal} with more examples, grouped by how many final decoder layers must be replaced before the prediction becomes correct. The baseline Qwen2.5-VL-3B-Instruct model initially generates incorrect answers for all examples presented in \Figref{fig:app-replace5ly} and \Figref{fig:app-replace20ly}. For the instances in \Figref{fig:app-replace5ly}, substituting only the final 5 decoder layers with those from the fine-tuned model is sufficient to correct the predictions. On the other hand, the examples in \Figref{fig:app-replace20ly} require a deeper layer replacement; substituting the final 20 layers is necessary before the model successfully recovers the correct answers.

\section{Per-Layer Semantic Readout Details}
\label{app:semantic-readout}

The readout probes are trained on the official Pixmo-Count training split~\citep{pixmo}, using only examples the frozen model answers correctly; no validation or test examples enter probe fitting. For each of the 36 decoder layers we train an independent three-layer MLP that maps the layer's final prompt-token state to a nine-class softmax distribution over the counts in \countrange{}. Splits are grouped by image hash to prevent near-duplicate leakage, probe outputs are temperature-calibrated, and results are averaged across five initialization seeds for Qwen-3B and three for each replication model. The retained correct-only pools contain 19,753, 21,705, and 19,317 examples for Qwen-3B, Qwen-7B, and LLaVA, respectively. Qwen numerals are single tokens; for LLaVA/Mistral we use the shared numeral-prefix position followed by the distinct digit token. The green curve reported in the main text is held-out probe accuracy on this split. Full optimizer, class-weighting, and calibration settings are shared with the paired analysis and are listed in Appendices~\ref{app:paired-bridge-details} and~\ref{app:cross-architecture-details}.

\begin{figure}[thb]
    \centering
    \begin{subfigure}[b]{0.3\linewidth}
        \centering
        \includegraphics[width=\linewidth]{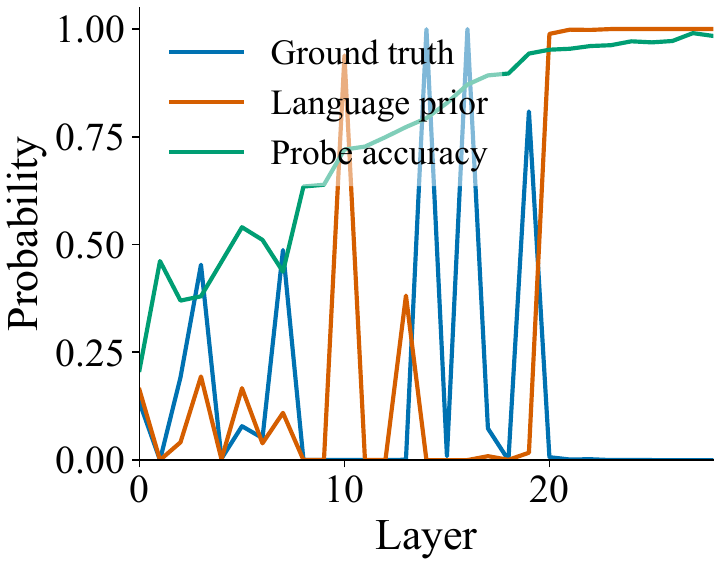}
        \caption{Example 1.}
        \label{fig:app-probing-example-1}
    \end{subfigure}
    \hspace{0.03\linewidth}
    \begin{subfigure}[b]{0.3\linewidth}
        \centering
        \includegraphics[width=\linewidth]{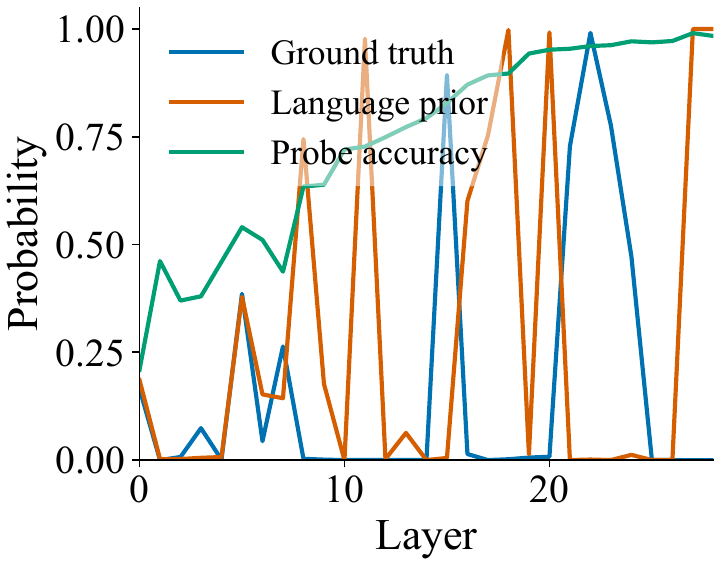}
        \caption{Example 2.}
        \label{fig:app-probing-example-2}
    \end{subfigure}
    \hspace{0.03\linewidth}
    \begin{subfigure}[b]{0.3\linewidth}
        \centering
        \includegraphics[width=\linewidth]{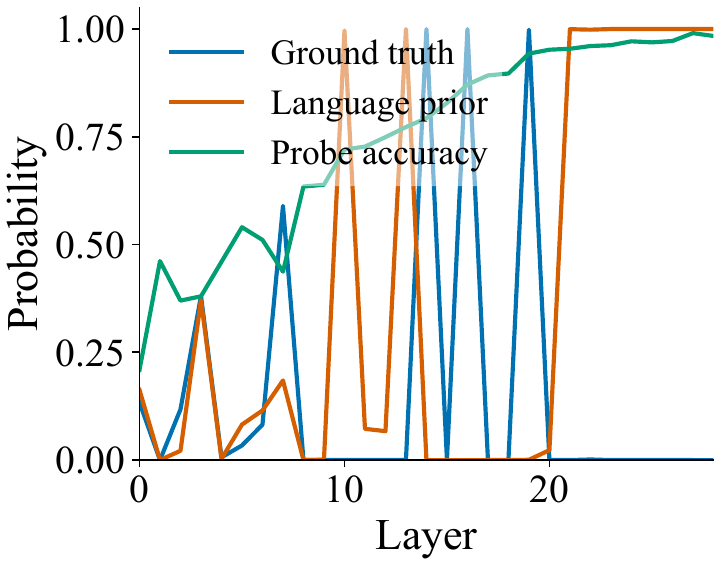}
        \caption{Example 3.}
        \label{fig:app-probing-example-3}
    \end{subfigure}

    \begin{subfigure}[b]{0.3\linewidth}
        \centering
        \includegraphics[width=\linewidth]{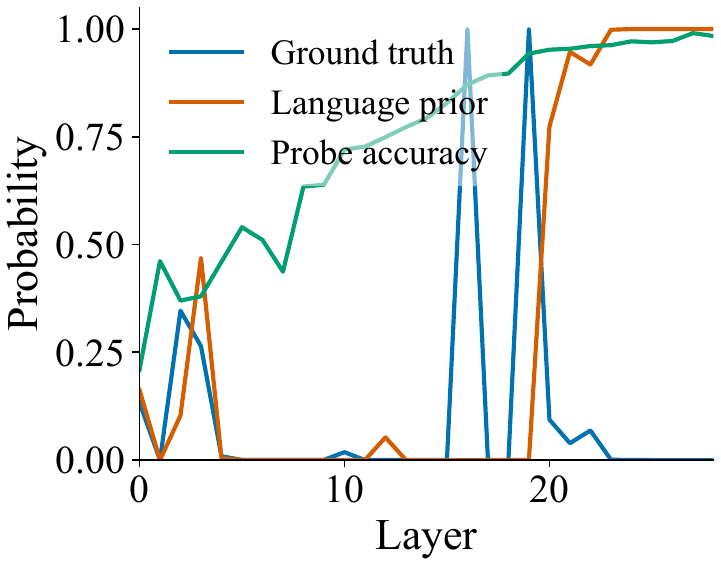}
        \caption{Example 4.}
        \label{fig:app-probing-example-4}
    \end{subfigure}
    \hspace{0.03\linewidth}
    \begin{subfigure}[b]{0.3\linewidth}
        \centering
        \includegraphics[width=\linewidth]{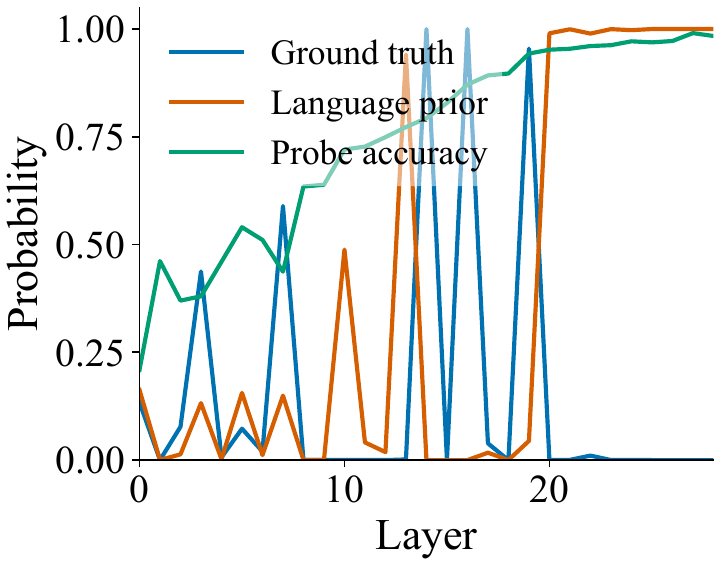}
        \caption{Example 5.}
        \label{fig:app-probing-example-5}
    \end{subfigure}
    \hspace{0.03\linewidth}
    \begin{subfigure}[b]{0.3\linewidth}
        \centering
        \includegraphics[width=\linewidth]{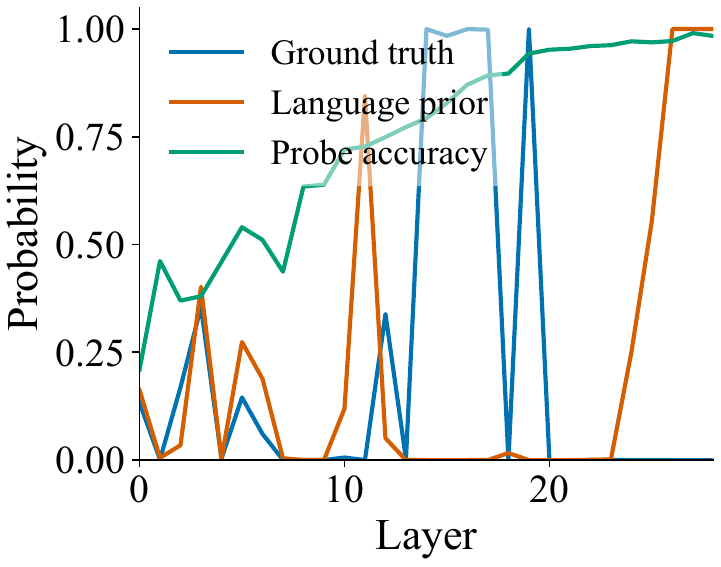}
        \caption{Example 6.}
        \label{fig:app-probing-example-6}
    \end{subfigure}

    \begin{subfigure}[b]{0.3\linewidth}
        \centering
        \includegraphics[width=\linewidth]{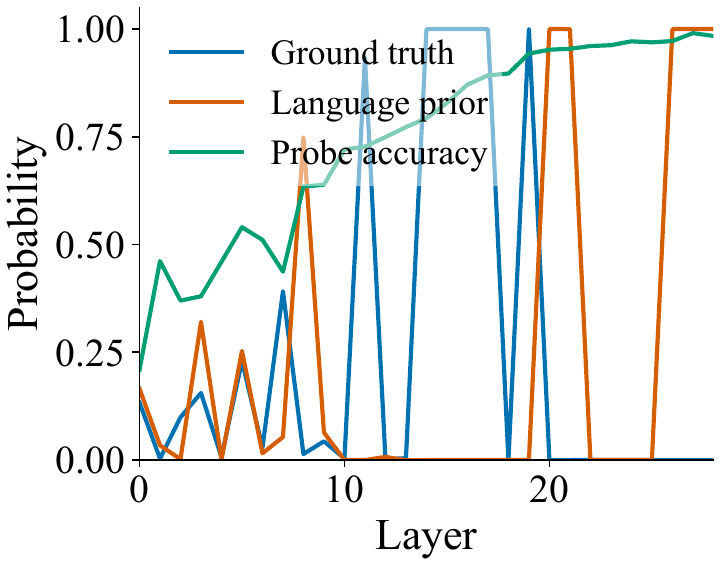}
        \caption{Example 7.}
        \label{fig:app-probing-example-7}
    \end{subfigure}
    \hspace{0.03\linewidth}
    \begin{subfigure}[b]{0.3\linewidth}
        \centering
        \includegraphics[width=\linewidth]{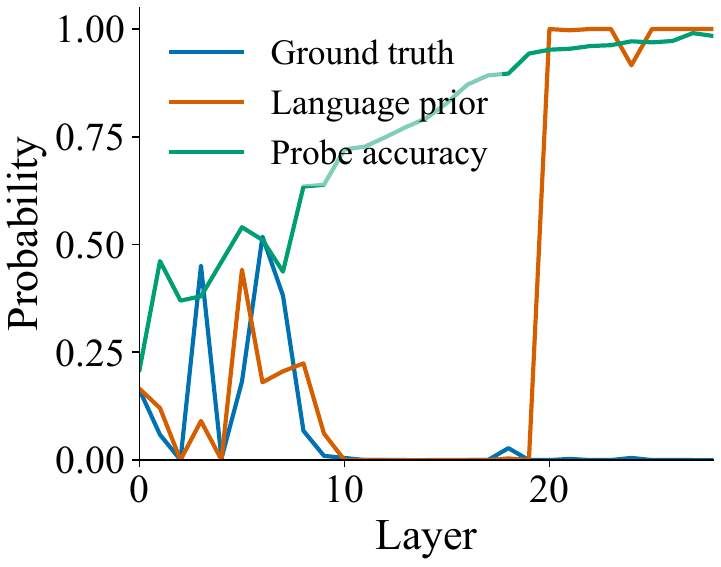}
        \caption{Example 8.}
        \label{fig:app-probing-example-8}
    \end{subfigure}
    \hspace{0.03\linewidth}
    \begin{subfigure}[b]{0.3\linewidth}
        \centering
        \includegraphics[width=\linewidth]{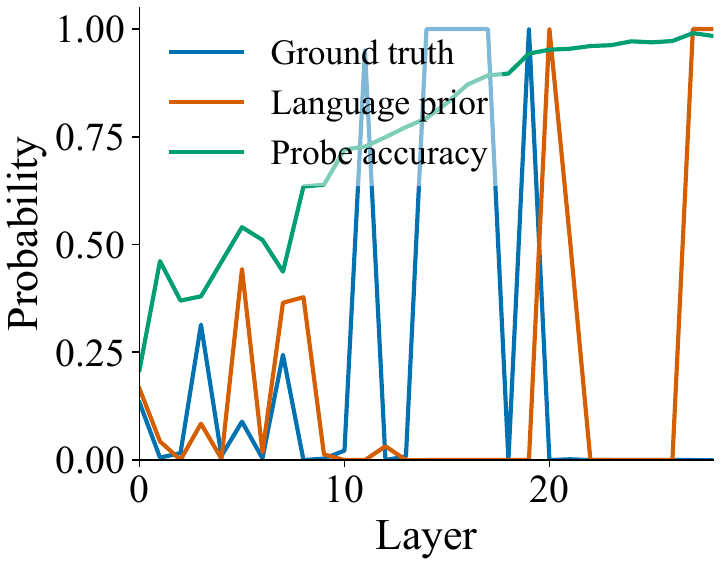}
        \caption{Example 9.}
        \label{fig:app-probing-example-9}
    \end{subfigure}

    \caption{Per-layer semantic readout on selected \qwensevenb{} VLMBias examples.}
    \label{fig:app-probing-more-examples}
\end{figure}

Beyond the population averages shown in the main text, \Figref{fig:app-probing-more-examples} shows the readout on individual \qwensevenb{} VLMBias examples. In most cases the ground-truth score temporarily exceeds the language-prior score at an intermediate layer before the late layers again favor the prior; one example instead ends on a third count. These cases illustrate highly non-monotonic probe trajectories, but they are qualitative examples rather than a population-level estimate of representational instability.

\FloatBarrier
\section{Paired-Bridge Experimental Details}
\label{app:paired-bridge-details}

This appendix gives the full setup behind the paired analysis in Section~\ref{sec:internal}: how the probes are trained, how the internal and output updates are defined, which reliability checks an effect must pass, and the diagnostic figures that support the main result.

\paragraph{Data and probes.} Both Pixmo-Count and paired VLMBias images use the model-native processor with visual-token bounds corresponding to $224^2$ minimum and $512^2$ maximum pixels. The official Pixmo-Count training split has 36,916 rows, of which 30,547 have counts in \countrange{}. Of these, 29,383 rows have locally recoverable images and 1,164 are unavailable; after removing 386 repeated image--label example IDs, we extract 28,997 unique examples. The frozen \qwenthreeb{} model answers 19,753 correctly, and only these examples enter probe fitting; no validation or test split from Pixmo-Count is used. Probe features are extracted with the prompt ``How many \{objects\} are there in the image? Answer with a single Arabic numeral, without any additional text.'' For paired VLMBias evaluation, Q1 preserves the source question form (for example, ``How many legs does this animal have?'') and Q2 uses its matched command paraphrase (``Count the legs of this animal.''); both receive the same single-numeral suffix. Analogous templates cover chess pieces, flag features, logo elements, and board rows or columns; optical-illusion pairs are excluded from the 197 numeric-count subset.

We train an independent three-layer MLP probe for each of the 36 decoder layers, with hidden widths 128 and 32, GELU activations, AdamW at learning rate $10^{-3}$ and weight decay $10^{-4}$, inverse-frequency class weights, and batch size 256. A single 70:15:15 split grouped by image hash is fixed with seed 2027; only initialization and minibatch order vary across seeds 2027--2031. We early-stop on the validation split and fit one temperature per layer there to calibrate the probe outputs. The retained counts for numerals 1 through 9 are 4,505, 3,921, 3,574, 3,386, 2,616, 930, 471, 297, and 53, so the largest counts are comparatively rare.

\paragraph{Internal and output updates.} For each pair $p$, $A_{\mathrm{GT}}$ is the counterfactual target count and $A_{\mathrm{LP}}$ is the normal-image reference count; both labels remain fixed across the two image conditions. In either condition, the probe margin $s_{p,l}$ is the temperature-calibrated probe logit for $A_{\mathrm{GT}}$ minus that for $A_{\mathrm{LP}}$, while the output margin $t_p$ uses the model's own output-head logits. The internal and output updates are $u_{p,l}=s_{p,l}^{\mathrm{cf}}-s_{p,l}^{\mathrm{normal}}$ and $M_p=t_p^{\mathrm{cf}}-t_p^{\mathrm{normal}}$, respectively; positive values indicate movement toward the counterfactual target. For each model and prompt, we fit
\[
\widehat u_{p,l}=\frac{u_{p,l}-\mu_l}{\sigma_l},\qquad
M_p=a_{g(p),l}+\beta_l\widehat u_{p,l}+\varepsilon_{p,l},
\]
where $\mu_l$ and $\sigma_l$ are the mean and population standard deviation of $u_{p,l}$ across the evaluated pairs, $g(p)$ denotes the pair's topic, $a_{g(p),l}$ is a topic-specific intercept, and $\varepsilon_{p,l}$ is the residual. The output update $M_p$ remains in logit units, so $\beta_l$ measures output-margin change per standard deviation of internal update. The main text uses $\beta$ for the corresponding layer or range coefficient.

\paragraph{Reliability checks.} We treat a layer's effect as reliable only if it passes all of the following: significance after multiple-comparison correction (Benjamini--Hochberg across layers, Holm across the predefined layer ranges), replication under the Q2 wording, agreement in four of five probe seeds, agreement in four of five leave-one-topic-out fits, adequate probe accuracy at that layer, a nonzero 95\% interval from a 10,000-sample topic-stratified bootstrap, and significance against a 10,000-permutation null that shuffles the output update within each topic. Every test uses the image pair as the sampling unit.

\paragraph{Layer ranges and sliding windows.} Panel (b) of \Figref{fig:paired-bridge} reports coupling for mean internal updates over layers 0--11, 12--23, 24--35, 28--35, and 32--35, for layer 35 alone, and for the late-minus-middle contrast (the mean over layers 24--35 minus that over 12--23). As a further check that the late effect is not an artifact of one particular range, we also slide a fixed-width window across depth and refit at every start layer, using widths of 2, 4, 6, and 12 layers. For each range, contrast, or window, we first form each pair's aggregated internal update, then standardize it across pairs and fit the same regression.

\begin{figure}[t]
    \centering
    \begin{subfigure}[b]{0.72\linewidth}
        \includegraphics[width=\linewidth]{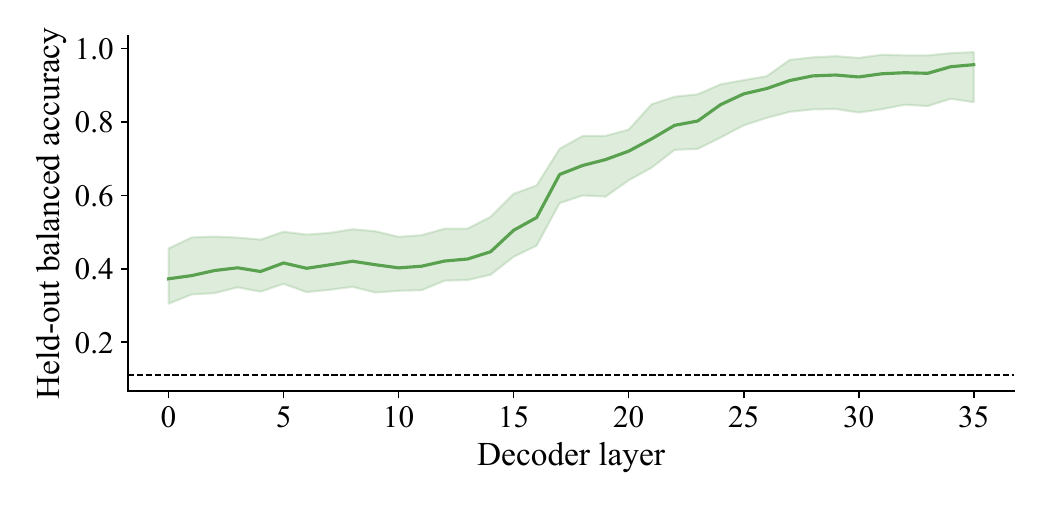}
        \caption{Probe accuracy on held-out examples, per layer.}
        \label{fig:paired-probe-quality}
    \end{subfigure}

    \vspace{0.6em}
    \begin{subfigure}[b]{0.72\linewidth}
        \includegraphics[width=\linewidth]{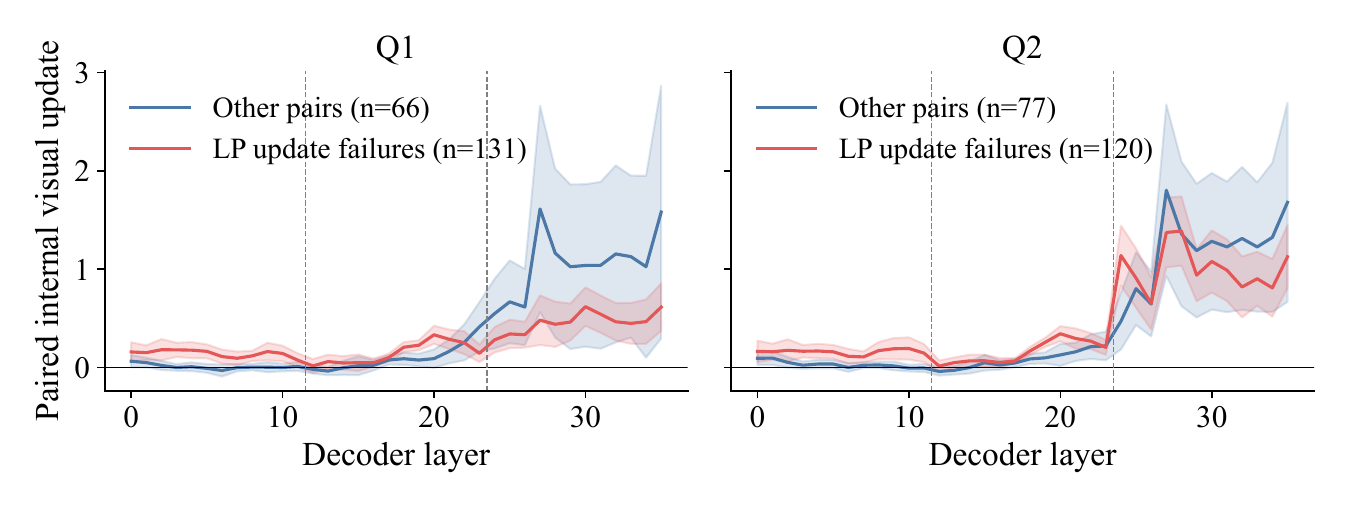}
        \caption{Mean internal update for LP update failures and other pairs.}
        \label{fig:paired-trajectories}
    \end{subfigure}
    \caption{Diagnostics for the paired analysis. (a) Shaded regions span bootstrap bounds across probe seeds. (b) Bands are 95\% confidence intervals around each group mean.}
    \label{fig:paired-diagnostics}
\end{figure}

\begin{figure}[t]
    \centering
    \includegraphics[width=0.9\linewidth]{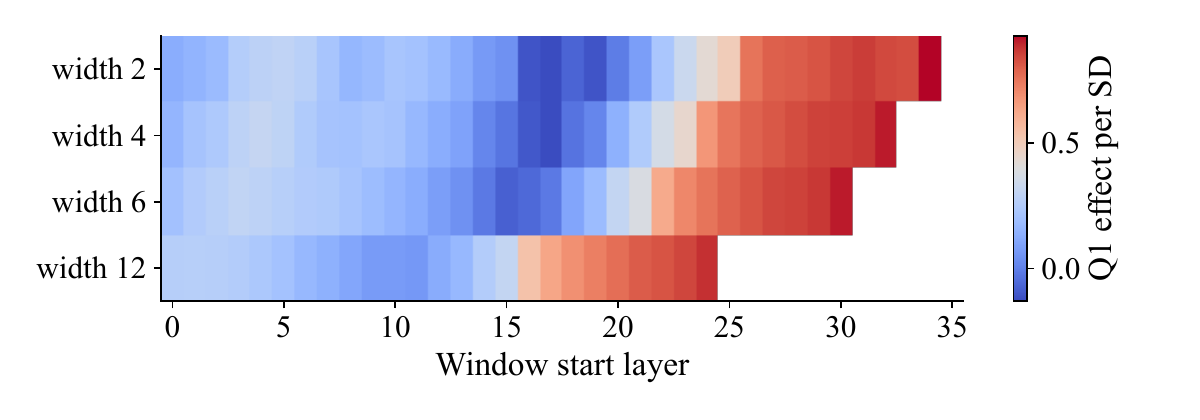}
    \caption{Coupling slope $\beta$ for sliding windows of widths 2, 4, 6, and 12 layers, plotted against the window's start layer. The late layers remain strong at every width, so the effect is not tied to one range definition.}
    \label{fig:paired-multiscale}
\end{figure}

\paragraph{Diagnostic figures.} \Figref{fig:paired-probe-quality} plots each layer's probe accuracy on the held-out test split (examples never seen during probe training); accuracy rises sharply after layer 16. \Figref{fig:paired-trajectories} splits the 197 pairs into LP update failures and other pairs, and plots the mean internal update across layers for each group; the two groups separate in the late layers. \Figref{fig:paired-multiscale} shows the slope $\beta$ for every sliding window; the late layers stay strong at all four window widths, so the result does not depend on one range definition.

\section{Activation Interchange Details}
\label{app:activation-interchange-details}

This appendix specifies the causal interchange experiment in Section~\ref{sec:causal-interchange}. We use all 197 Q1 pairs and run \qwenthreeb{} on both the normal and counterfactual images under the same prompt. The paired images use identical $768\times768$ canvases, yielding the same input length, attention mask, and final prompt-token position. All image tokens and preceding text tokens occur before this final prompt token in the causal sequence. This alignment lets us replace its state without changing the remaining sequence.

\paragraph{Residual interchange.}
We index the $L$ decoder blocks by $l=0,\ldots,L-1$. For pair $p$, let $H_{p,l}^{\mathrm{normal}}\in\mathbb{R}^{T\times d}$ be the input sequence to block $l$ for the normal image, where $T$ is the sequence length and $d$ is the hidden width. Writing $F_l$ for decoder block $l$, its output is
\[
H_{p,l+1}^{\mathrm{normal}}=F_l\!\left(H_{p,l}^{\mathrm{normal}}\right).
\]
An output intervention labeled $l$ replaces row $f$, the final prompt-token state, in $H_{p,l+1}^{\mathrm{normal}}$ with its counterfactual counterpart:
\[
\widetilde H_{p,l+1}^{\,\mathrm{normal}\leftarrow\mathrm{cf}}[j]
=
\begin{cases}
H_{p,l+1}^{\mathrm{cf}}[f], & j=f,\\
H_{p,l+1}^{\mathrm{normal}}[j], & j\neq f.
\end{cases}
\]
All other token states remain from the normal run at the intervention point. Because the image and preceding text tokens occur before position $f$, causal attention prevents the replacement at $f$ from changing their subsequent states. The modified final prompt token can still attend to these preceding states as the remaining decoder layers operate normally:
\[
\widetilde H_{p,k+1}^{\,\mathrm{normal}\leftarrow\mathrm{cf}}
=F_k\!\left(\widetilde H_{p,k}^{\,\mathrm{normal}\leftarrow\mathrm{cf}}\right),
\qquad k=l+1,\ldots,L-1.
\]
We repeat this intervention independently at each of the 36 layers and read the resulting logits for counts 1--9. We use the same GT-minus-LP output margin $t_p$ as in Section~\ref{sec:paired-counterfactual-updates}. Let $\widetilde t_{p,l}^{\mathrm{normal}\leftarrow\mathrm{cf}}$ denote this margin after replacing $H_{p,l+1}^{\mathrm{normal}}[f]$ with its counterfactual counterpart. The logit-margin update is
\[
\Delta t_{p,l}
=\widetilde t_{p,l}^{\mathrm{normal}\leftarrow\mathrm{cf}}-t_p^{\mathrm{normal}}.
\]
Positive values indicate movement toward GT. As an implementation check, replacing a state with itself produces exactly zero update at every layer.

For significance, we form 95\% confidence intervals with a 10,000-sample bootstrap over pairs and test each layer with a Wilcoxon signed-rank test, correcting across all 36 layers with the Holm method. Besides single layers, we also replace the normal states with their counterfactual counterparts across four fixed layer ranges (4--10, 12--18, 25--31, and 32--35) to summarize several layers at once. Finally, as a sensitivity check, we repeat the intervention using an unrelated image matched on prompt, topic, and both counts, for 12 pairs at six representative layers.

\section{Locating the Layer-27 Jump Within the Layer}
\label{app:block27-case-study}

This appendix gives the details behind Section~\ref{sec:internal}: how we test which step inside a decoder layer produces the sharp jump at layer 27, and the checks that confirm the interventions are implemented correctly.

Qwen-3B uses a pre-norm Transformer decoder. For a fixed pair and image condition, let $x=H_{p,l}$ denote the full input sequence to block $l$, $y$ its post-attention sequence, and $z=H_{p,l+1}$ its output sequence. The two residual updates are
\[
y=x+\operatorname{Attn}\!\left(\operatorname{RMSNorm}_{\mathrm{attn}}(x)\right),
\qquad
z=y+\operatorname{MLP}\!\left(\operatorname{RMSNorm}_{\mathrm{mlp}}(y)\right),
\]
where $\operatorname{RMSNorm}_{\mathrm{attn}}$ and $\operatorname{RMSNorm}_{\mathrm{mlp}}$ denote the separate normalization operations before the attention and MLP submodules. The states $x$, $y$, and $z$ are the three replacement points reported in Table~\ref{tab:block27-boundaries}. Because layers are indexed from zero, the increase between the preceding layer output and the layer-27 output is produced within the layer labeled 27.

At each boundary, we replace only row $f$, the normal run's final prompt-token state, with its counterfactual counterpart and recompute the remaining operations. We apply this intervention to all 197 Q1 pairs in layers 26, 27, and 28, using the same logit-margin update $\Delta t$ as in Section~\ref{sec:causal-interchange}. For each pair, attention gain is the post-attention $\Delta t$ minus the input $\Delta t$, and MLP gain is the output $\Delta t$ minus the post-attention $\Delta t$. Replacing the normal state with itself yields zero update, and the layer-input and layer-output replacements reproduce the corresponding points from the layerwise scan. Confidence intervals use the image pair as the sampling unit, and Holm correction is applied across the nine layer-by-step tests.

\section{Cross-Architecture Paired Analysis}
\label{app:cross-architecture-details}

The Qwen-7B and LLaVA replications use the same 28,997 unique Pixmo-Count training examples and all 788 paired records (197 pairs, two prompt variants, and two image conditions). Full artifact checks require unique IDs, finite tensors, and feature shapes $[28,3584]$ for \qwensevenb{} and $[32,4096]$ for \llava{}. The correct-only probe pools contain 21,705 and 19,317 examples, respectively. Three initialization seeds (2027--2029) share an image-hash-grouped 70:15:15 split, normalization statistics, held-out IDs, and calibration protocol; only initialization and minibatch order vary.

Qwen numerals are represented by nine distinct single tokens. Mistral tokenizes a bare numeral as a shared prefix token (ID 28705) followed by one of nine digit tokens. We append the shared prefix to each LLaVA prompt and extract the final-token representation and logits for the following digit decision. The LLaVA paired input contains 2,928 image tokens and has identical sequence geometry across pair members. For both architectures, a hook-based extractor that stores only the final-token state at every layer is compared against the full \texttt{output\_hidden\_states=True} path; features and output logits match bit-for-bit on paired smoke tests.

At each layer, we fit the topic-fixed-effect regression using the calibrated probe margins and predictor standardization defined in Appendix~\ref{app:paired-bridge-details}. Q1 is primary and Q2 is a wording replication. Layer-wise Q1 tests use Benjamini--Hochberg correction; 10,000-sample topic-stratified bootstrap intervals quantify uncertainty, and all three seeds must agree in direction. For $L$ decoder blocks indexed by $l=0,\ldots,L-1$, normalized depth is $\delta_l=l/(L-1)$, with $L=36,28,32$ for Qwen-3B, Qwen-7B, and LLaVA, respectively. We report this depth and predefined early, middle, late, final-quarter, and final-eighth intervals. The terminal contiguous band begins at the first layer after which every remaining layer meets the cross-prompt criteria, while the complete curve retains any earlier isolated effects. Partial $R^2$ is the fraction of residual variance explained by the internal update after controlling for topic. Its square root, signed by $\beta$, is correlation-like.

\section{Behavioral Metric Details and Representative Results}
\label{app:behavioral-details}

This appendix gives the exact definitions behind the four metrics reported in Section~\ref{sec:behavioral} and shows the per-benchmark trajectories omitted from the main text. For each sample $(I,Q,\tilde A)$, where $\tilde A$ is the ground-truth answer, and each blur level $k_i$, $E_{I,Q,k_i}$ indicates single-level agreement, meaning that the current answer matches the original-image answer. The indicator $C_{I,Q,k_i}$ records consecutive agreement, meaning that this match has held at every level from $k_1$ through $k_i$. Averaging these indicators over a dataset gives
\[
\dseqprob=\E_{\ds}[E_{I,Q,k_i}],\qquad
\dsconsprob=\E_{\ds}[C_{I,Q,k_i}],
\]
the single-level and consecutive agreement rates, respectively.

Overall accuracy at blur level $k_i$ is $\dsbluracc=\E_{\ds}[\mathbb{1}(A_{I,Q,k_i}=\tilde A)]$. We also report invariant-subset accuracy over $\ds_{k_i}=\{(I,Q,\tilde A)\in\ds:C_{I,Q,k_i}=1\}$, the subset whose answers have remained unchanged through $k_i$. For a nonempty subset,
\[
\dsconsacc=\E_{(I,Q,\tilde A)\sim\ds_{k_i}}\left[\mathbb{1}(A_{I,Q,k_i}=\tilde A)\right],
\]
where the expectation is the uniform empirical mean over $\ds_{k_i}$. Because the invariant subset becomes smaller as blur increases, its accuracy can exceed the unblurred full-dataset accuracy. A high invariant-subset accuracy under severe blur means that the benchmark continues to reward many predictions that do not respond to the removed visual detail.

All blur conditions within a benchmark use the same evaluation samples, prompt format, greedy decoding, and answer-extraction rule; only the image changes. Greedy decoding ensures that changes along the blur path reflect the image intervention rather than sampling variability. Consecutive agreement therefore measures the stability of each deterministic prediction across the tested blur sequence. Dataset-specific prompting and answer normalization follow each benchmark's evaluator.

\section{Visual-Decay Controls Across Benchmarks}
\label{app:visual-decay-controls}

We complement progressive Gaussian blur with visual-decay controls across the same twelve benchmarks and three models. Grayscale removes color while preserving luminance and geometry. For nested spatial masking, we divide each image into an $8\times8$ grid and replace a deterministic nested subset of 25\%, 50\%, or 75\% of cells with neutral gray. Thus, cells masked at 25\% remain masked at 50\% and 75\%. The text-only control removes image content from the prompt. These controls test whether the blur results also appear when visual evidence is reduced through color removal, spatial occlusion, or complete removal.

\begin{table}[t]
\centering
\caption{Benchmark-scale visual-decay controls across \qwenthreebshort{}, \qwensevenbshort{}, and \llavashort{}. All values are percentages and micro-averages over the same 42,078 examples from 12 benchmarks. ``Same'' is exact agreement with the corresponding clean-image answer. Nested masks divide each image into an $8\times8$ grid and replace nested 25\%, 50\%, and 75\% subsets of cells with neutral gray; ``all-level same'' requires agreement under the original image and all three mask levels, while its paired accuracy is measured within that unchanged subset. Blur values use the historical blur-run clean answer; grayscale, masks, and text-only use the fresh clean answer generated with the same inference stack.}
\label{tab:visual-decay-controls}
\footnotesize
\setlength{\tabcolsep}{5pt}
\renewcommand{\arraystretch}{1.25}
\begin{tabular}{l c c c c c}
\hline
\textbf{Model} & \textbf{Clean acc.} & \textbf{Blur} $\ksz{61}$ & \textbf{Grayscale} & \textbf{Nested masks} & \textbf{Text-only} \\
& & acc. / same & acc. / same & all-level same / acc. & acc. / same \\
\hline
        Qwen-3B & 65.0 & 37.2 / 46.9 & 61.1 / 79.6 & 48.1 / 74.4 & 34.8 / 43.8 \\
        Qwen-7B & 68.1 & 34.1 / 42.6 & 63.7 / 83.4 & 53.4 / 76.6 & 37.8 / 45.1 \\
        LLaVA-7B & 54.7 & 34.8 / 53.5 & 51.8 / 89.6 & 65.0 / 57.6 & 34.2 / 46.9 \\
\hline
\end{tabular}
\end{table}

The masking results show the same broad pattern as blur. The fractions of predictions unchanged from the original image through all three mask levels are 48.1\%, 53.4\%, and 65.0\% for \qwenthreebshort{}, \qwensevenbshort{}, and \llavashort{}, respectively. Their accuracy within these unchanged subsets is 74.4\%, 76.6\%, and 57.6\%. Text-only accuracy remains 34.8\%, 37.8\%, and 34.2\%, indicating that benchmark credit can persist without image input. These controls do not show that every unchanged answer ignores all visual evidence: grayscale and masks retain some cues, and a stable answer can remain justified when sufficient evidence survives. Instead, together with progressive blur, they show that benchmark accuracy alone is incomplete evidence that a model uses the visual detail affected by the tested visual decay.

\section{Per-Benchmark Blur Trajectories and More Qualitative Examples}
\label{app:more-datasets}

\begin{figure}[t]
    \centering
    \begin{subfigure}[b]{0.95\linewidth}
        \includegraphics[width=\linewidth]{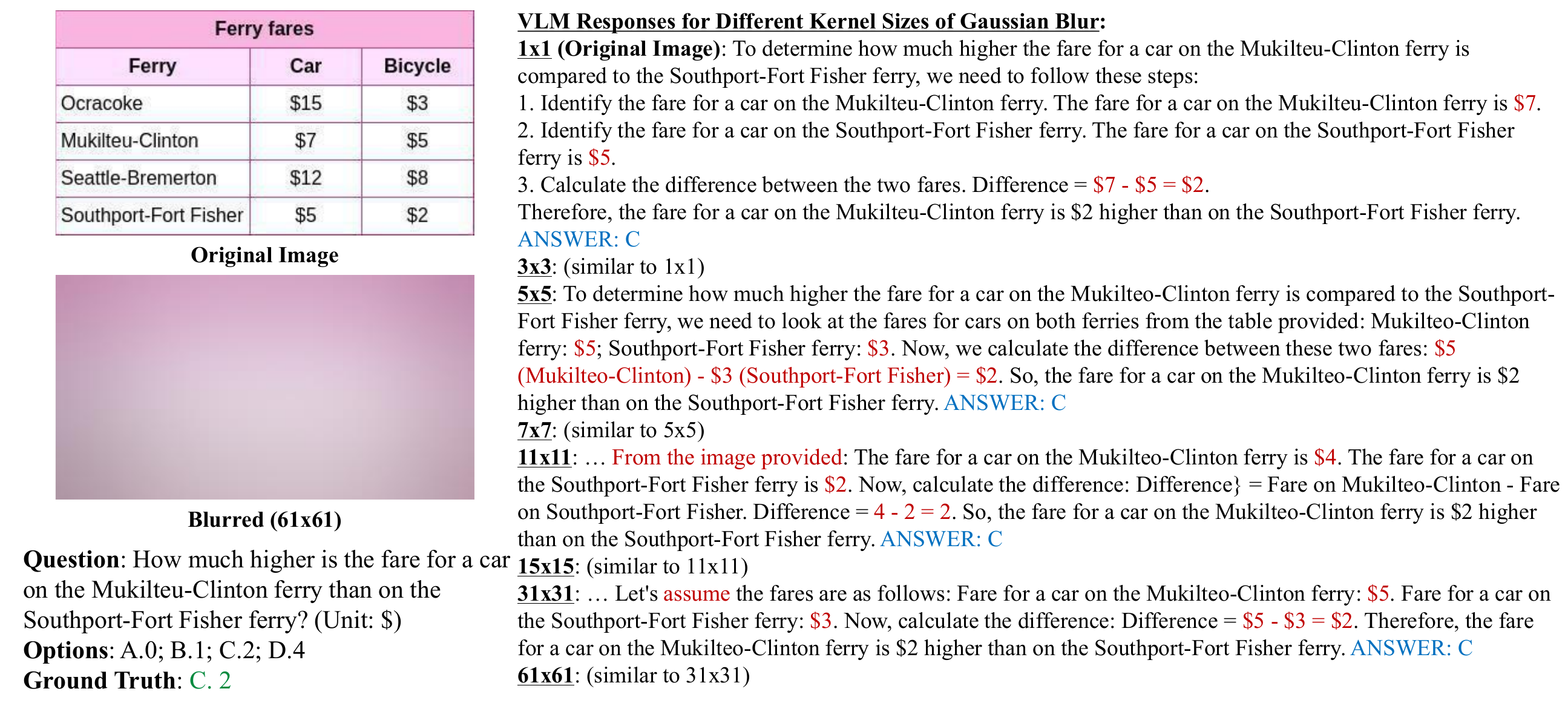}
        \caption{Invariant and correct.}
        \label{fig:app-langprior-right-example}
    \end{subfigure}
    \caption{A correct invariant MMStar response from \qwensevenb{}. As blur increases, changing intermediate OCR values are rationalized into the same correct difference. The complementary incorrect example appears in \Figref{fig:langprior-example-main}.}
    \label{fig:app-langprior-examples}
\end{figure}

\Figref{fig:app-dataset-results} collects the per-benchmark trajectories, with one row for each benchmark or diagnostic partition and one curve per model. The dataset-averaged trajectories in \Figref{fig:model-comparison} are unweighted means over the twelve benchmarks listed in Section~\ref{sec:behavioral}. HallusionBench Visual Dependent and Visual Supplement appear below as diagnostic partitions and do not enter the average as additional benchmarks. A MMStar example in which \qwensevenbshort{} gives the correct answer and remains invariant through all blur levels follows the complete figure series in \Figref{fig:app-langprior-examples}.

\begin{figure}[t]
    \centering
    \begin{subfigure}[b]{0.95\linewidth}
        \includegraphics[width=\linewidth,trim=0 22bp 0 18bp,clip]{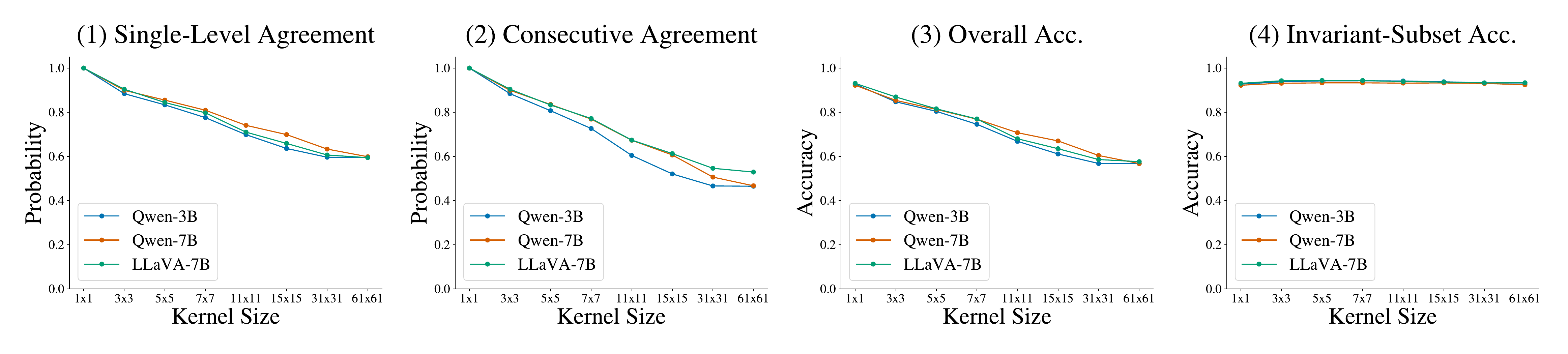}
        \caption{RePOPE~\citep{repope}.}
        \label{fig:ds-repope}
    \end{subfigure}
    \begin{subfigure}[b]{0.95\linewidth}
        \includegraphics[width=\linewidth]{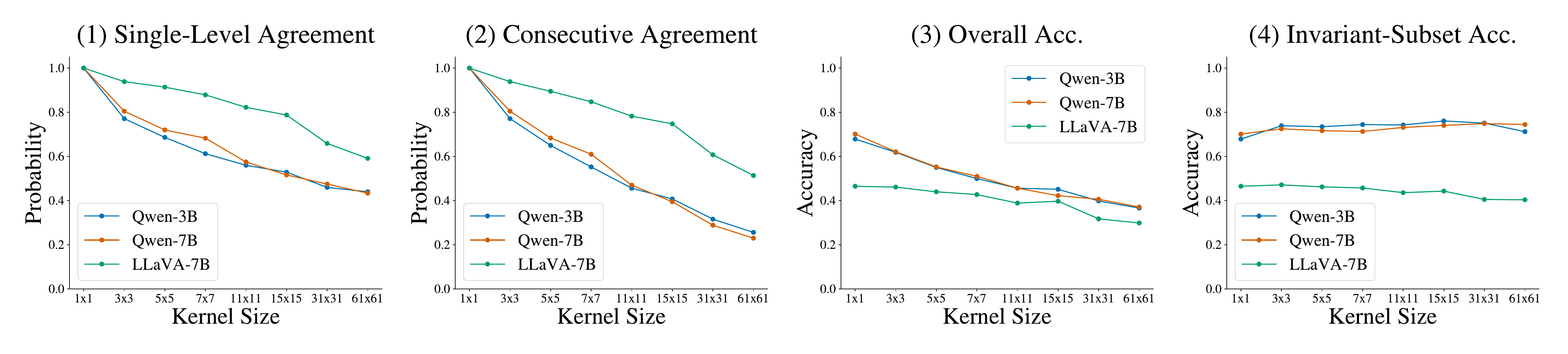}
        \caption{HR-Bench~\citep{hrbench}.}
        \label{fig:app-ds-hrbench}
    \end{subfigure}
    \begin{subfigure}[b]{0.95\linewidth}
        \includegraphics[width=\linewidth,trim=0 22bp 0 18bp,clip]{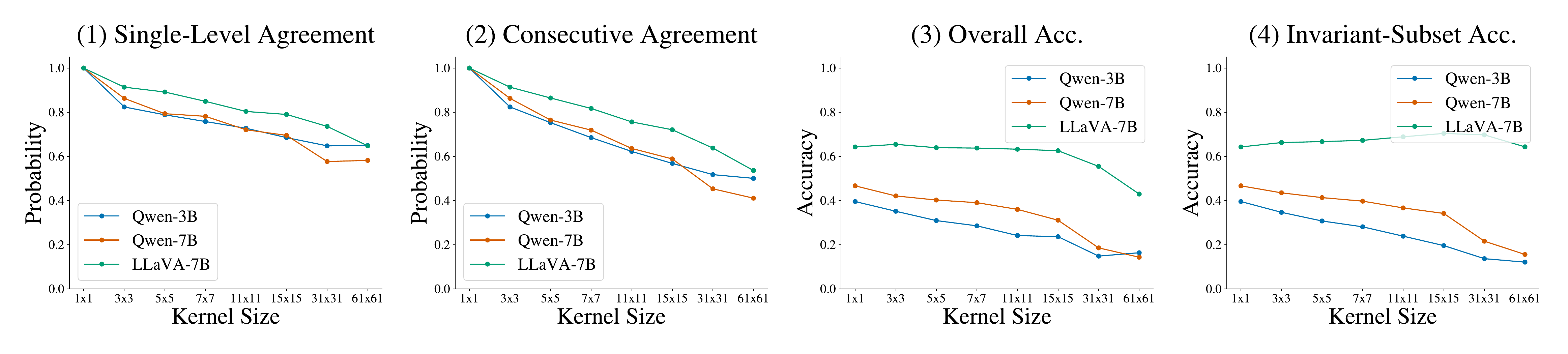}
        \caption{HallusionBench Visual Dependent~\citep{hallusionbench}.}
        \label{fig:ds-hallusionbench-vd}
    \end{subfigure}
    \begin{subfigure}[b]{0.95\linewidth}
        \includegraphics[width=\linewidth]{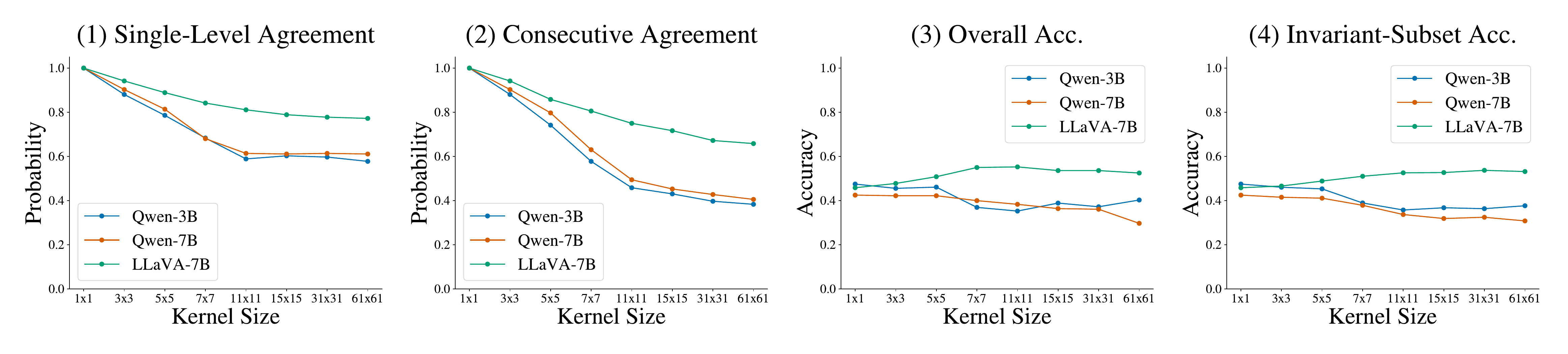}
        \caption{HallusionBench Visual Supplement~\citep{hallusionbench}.}
        \label{fig:app-ds-hallusionbench-vs}
    \end{subfigure}
    \begin{subfigure}[b]{0.95\linewidth}
        \includegraphics[width=\linewidth]{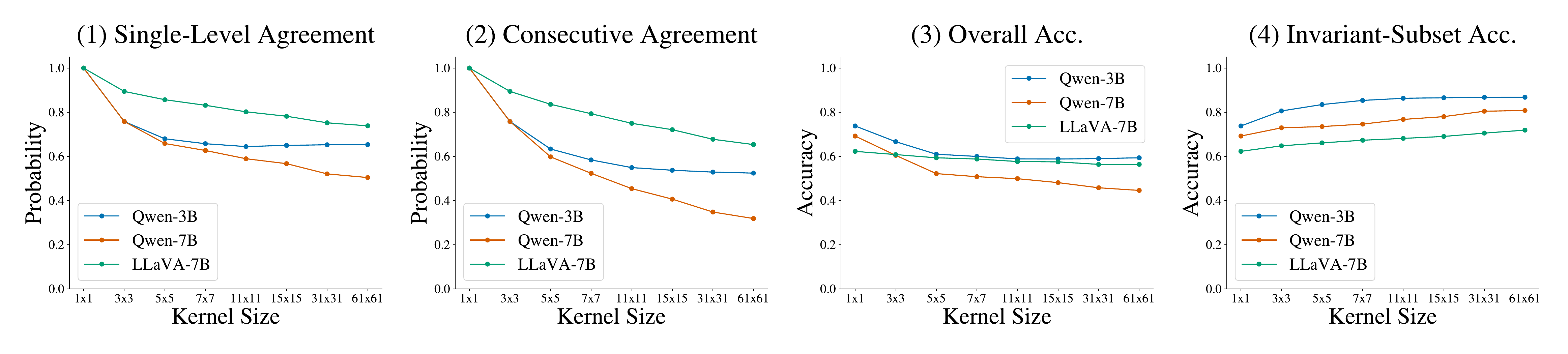}
        \caption{AI2D~\citep{ai2d}.}
        \label{fig:ds-appendix-ai2d}
    \end{subfigure}
    \caption{Per-benchmark trajectories under increasing Gaussian blur for \qwenthreebshort{}, \qwensevenbshort{}, and \llavashort{}. From left to right, the panels show single-level agreement $\dseqprob$, where the current answer matches the original-image answer; consecutive agreement $\dsconsprob$, where that match holds at every level through the current level; overall accuracy $\dsbluracc$; and invariant-subset accuracy $\dsconsacc$. Panels 3 and 4 report raw accuracies, and each curve represents one model.}
    \label{fig:app-dataset-results}
\end{figure}

\begin{figure}[t]
    \ContinuedFloat
    \centering
    \begin{subfigure}[b]{0.95\linewidth}
        \includegraphics[width=\linewidth]{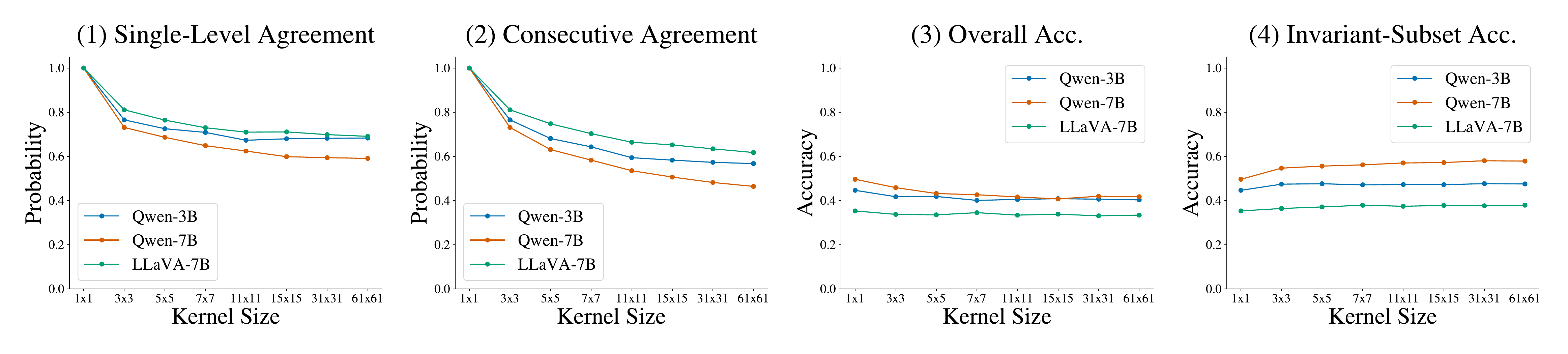}
        \caption{MMMU~\citep{mmmu}.}
        \label{fig:ds-appendix-mmmu}
    \end{subfigure}
    \begin{subfigure}[b]{0.95\linewidth}
        \includegraphics[width=\linewidth]{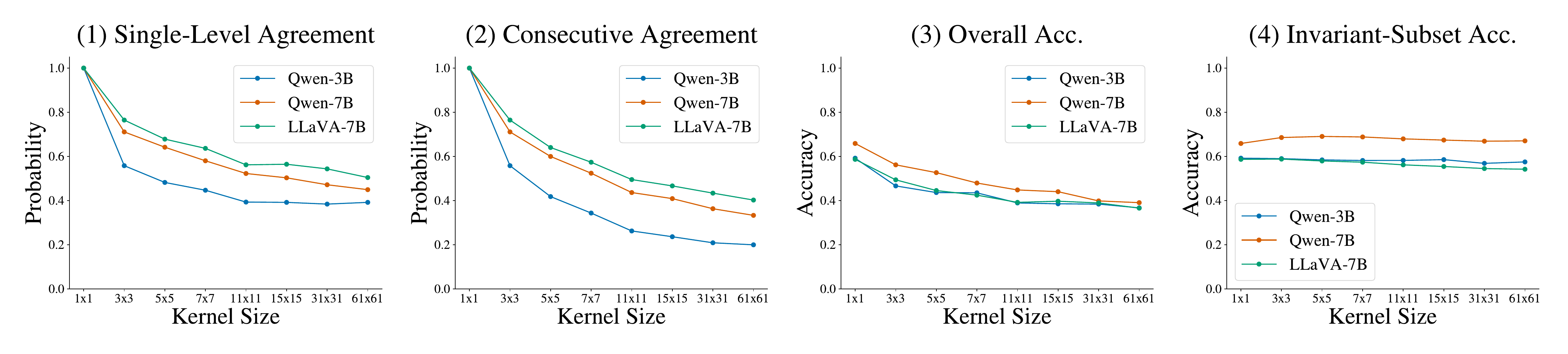}
        \caption{RealworldQA~\citep{realworldqa}.}
        \label{fig:ds-appendix-realworldqa}
    \end{subfigure}
    \begin{subfigure}[b]{0.95\linewidth}
        \includegraphics[width=\linewidth]{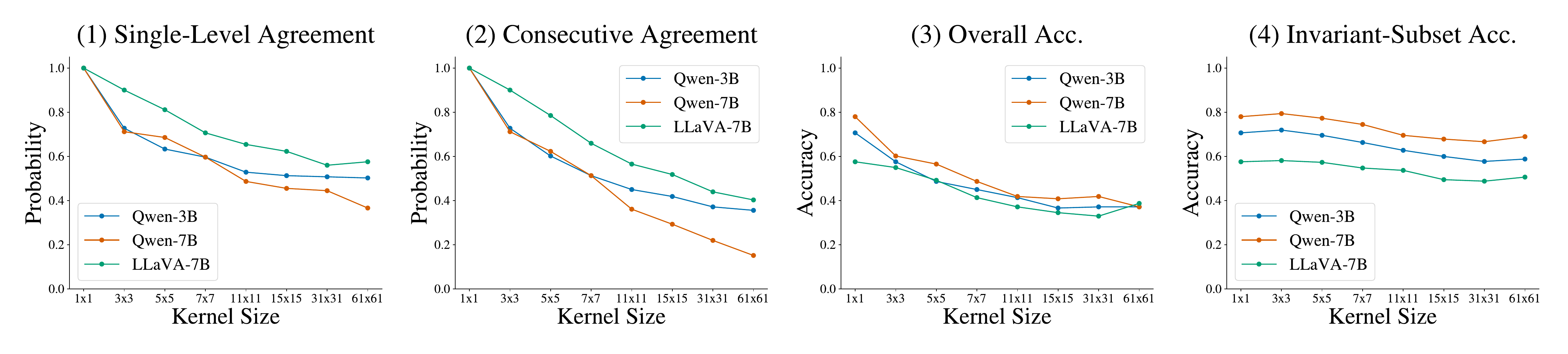}
        \caption{V*Bench~\citep{vstar}.}
        \label{fig:ds-appendix-vstar}
    \end{subfigure}
    \begin{subfigure}[b]{0.95\linewidth}
        \includegraphics[width=\linewidth]{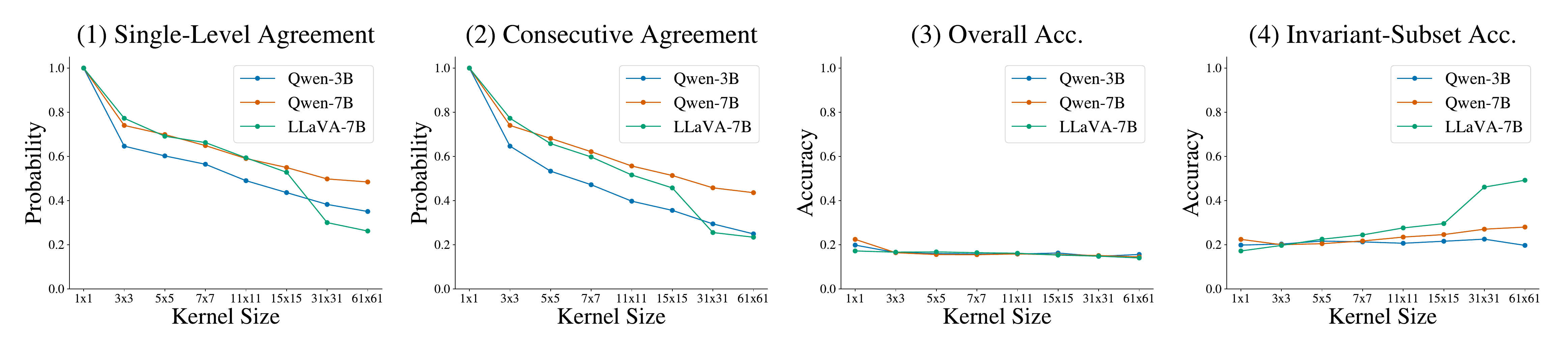}
        \caption{VLMBias~\citep{vlmbias}.}
        \label{fig:ds-appendix-vlmbias}
    \end{subfigure}
    \begin{subfigure}[b]{0.95\linewidth}
        \includegraphics[width=\linewidth]{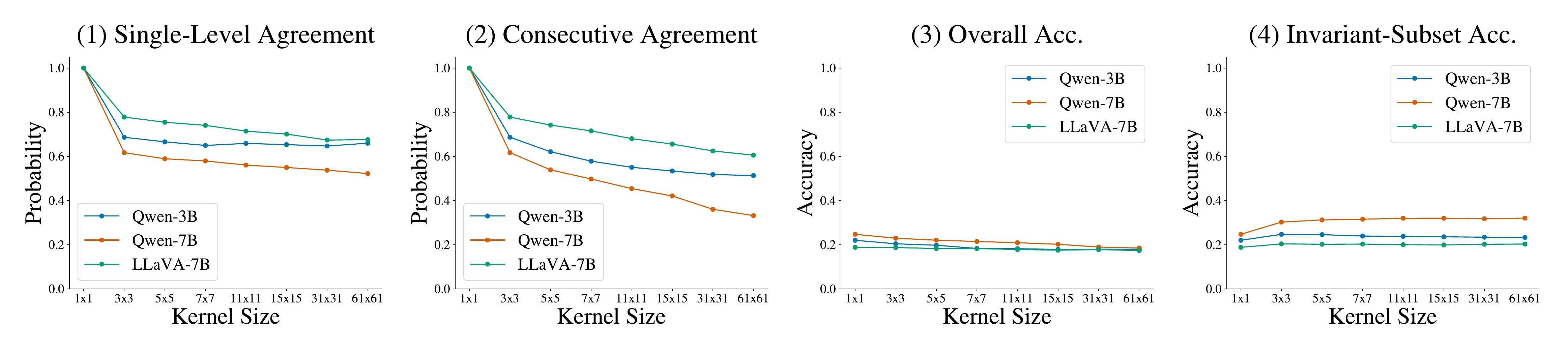}
        \caption{BLINK~\citep{blink}.}
        \label{fig:ds-appendix-blink}
    \end{subfigure}
    \caption{Per-benchmark trajectories under increasing Gaussian blur for \qwenthreebshort{}, \qwensevenbshort{}, and \llavashort{} (continued). From left to right, the panels show single-level agreement $\dseqprob$, consecutive agreement $\dsconsprob$, raw overall accuracy $\dsbluracc$, and raw invariant-subset accuracy $\dsconsacc$. Each curve represents one model.}
\end{figure}

\begin{figure}[t]
    \ContinuedFloat
    \centering
    \begin{subfigure}[b]{0.95\linewidth}
        \includegraphics[width=\linewidth]{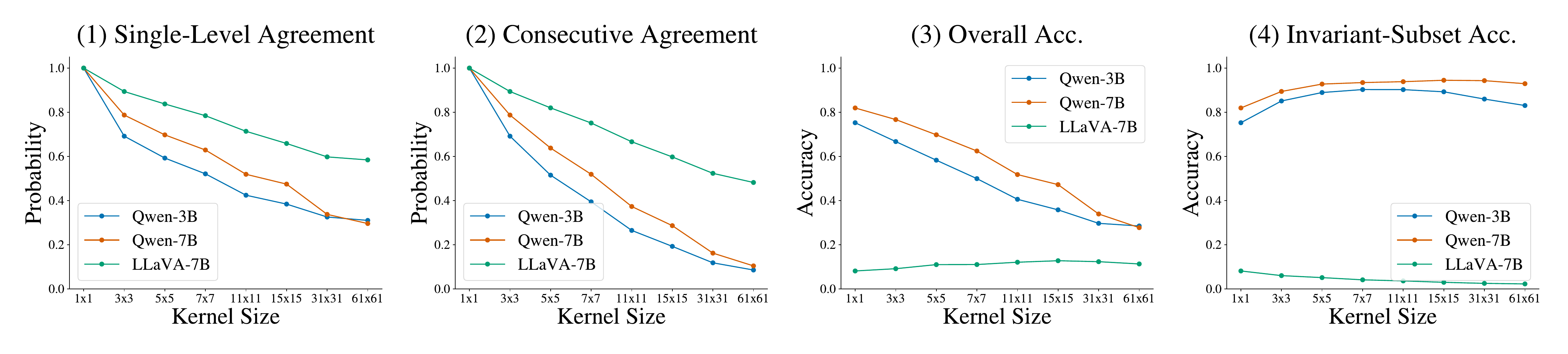}
        \caption{MMBench~\citep{mmbench}.}
        \label{fig:ds-appendix-mmbench}
    \end{subfigure}
    \begin{subfigure}[b]{0.95\linewidth}
        \includegraphics[width=\linewidth]{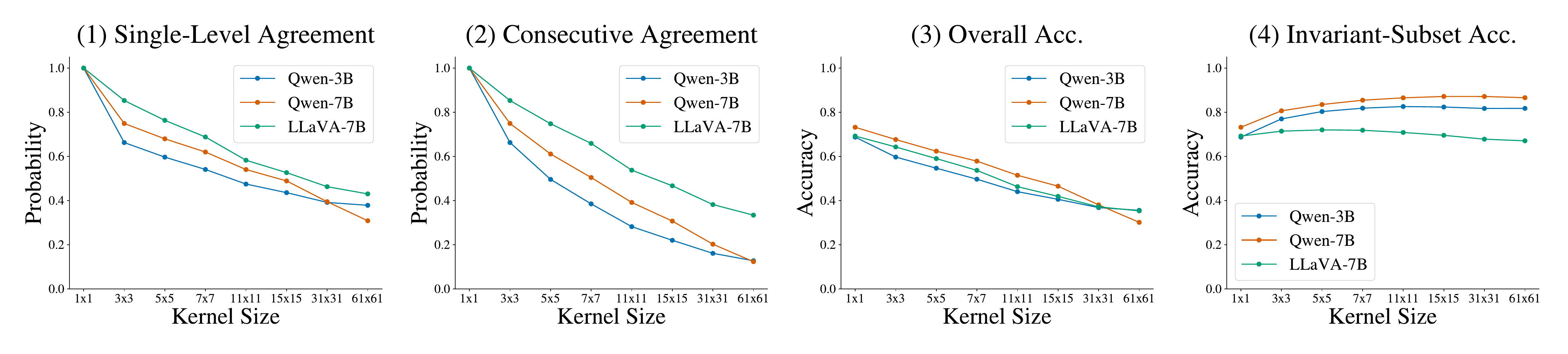}
        \caption{SEED-Bench~\citep{seedbench}.}
        \label{fig:ds-appendix-seedbench}
    \end{subfigure}
    \begin{subfigure}[b]{0.95\linewidth}
        \includegraphics[width=\linewidth]{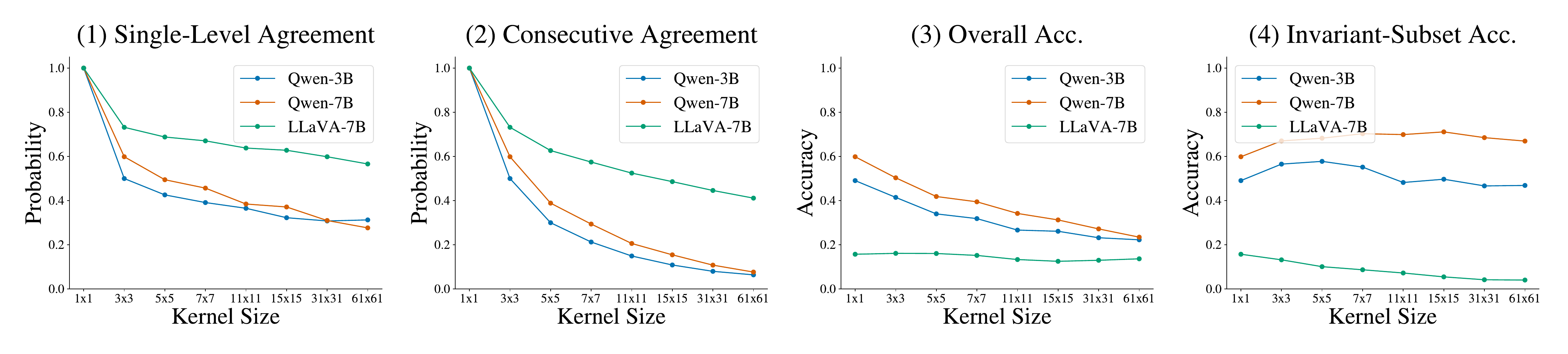}
        \caption{MMStar~\citep{mmstar}.}
        \label{fig:ds-appendix-mmstar}
    \end{subfigure}
    \caption{Per-benchmark trajectories under increasing Gaussian blur for \qwenthreebshort{}, \qwensevenbshort{}, and \llavashort{} (continued). From left to right, the panels show single-level agreement $\dseqprob$, consecutive agreement $\dsconsprob$, raw overall accuracy $\dsbluracc$, and raw invariant-subset accuracy $\dsconsacc$. Each curve represents one model.}
\end{figure}

\end{document}